\documentclass{article}

 \usepackage[preprint]{neurips_2026}

\usepackage[utf8]{inputenc} % allow utf-8 input
\usepackage[T1]{fontenc}    % use 8-bit T1 fonts
\usepackage{hyperref}       % hyperlinks
\usepackage{url}            % simple URL typesetting
\usepackage{booktabs}       % professional-quality tables
\usepackage{amsfonts}       % blackboard math symbols
\usepackage{nicefrac}       % compact symbols for 1/2, etc.
\usepackage{microtype}      % microtypography
\usepackage{xcolor}         % colors

\usepackage{microtype}
\usepackage{graphicx}
\usepackage{subcaption}
\usepackage{booktabs} % for professional tables
\usepackage{pifont}
\usepackage{hyperref}
\usepackage{bm}
\usepackage{wrapfig}
\usepackage{amsmath}
\usepackage{amssymb}
\usepackage{mathtools}
\usepackage{amsthm}
\usepackage{multicol}
\usepackage{multirow}
\usepackage{comment}
\usepackage{enumitem}
\usepackage{marvosym}
\usepackage[most]{tcolorbox}
\tcbset{
  promptbox/.style={
    enhanced,
    breakable,
    colframe=brown!50!black,
    colback=brown!4,
    boxrule=0.4pt,
    arc=2pt,
    left=6pt,right=6pt,top=6pt,bottom=6pt,
    fonttitle=\bfseries,
    fontupper=\normalfont\small,
    title={#1},
  }
}

\definecolor{bluecite}{HTML}{0875b7}
\hypersetup{
  colorlinks=true,
  citecolor=bluecite,
  linkcolor=magenta
}

\title{InsightTok: Improving Text and Face Fidelity\\in Discrete Tokenization\\for Autoregressive Image Generation}
\newcommand{\methodname}{InsightTok}

\author{
Yang Yue$^\dagger$~~~~Fangyun Wei$^\ddagger$~~~~Tianyu He$^\ddagger$~~~~Jinjing Zhao$^\ddagger$~~~~Zanlin Ni$^\dagger$~~~~Zeyu Liu$^\dagger$\\ 
\textbf{Jiayi Guo$^\dagger$~~~~Lei Shi$^\ddagger$~~~~Yue Dong$^\ddagger$~~~~Li Chen$^\ddagger$~~~~Ji Li$^\ddagger$~~~~Gao Huang$^{\dagger,\text{\Letter}}$~~~~Dong Chen$^{\ddagger,\text{\Letter}}$}\\
$^\dagger$Tsinghua University~~~~~~$^\ddagger$Microsoft Research\\
\\
\url{https://github.com/LeapLabTHU/InsightTok}
}

\begin{document}

\begingroup
\renewcommand\thefootnote{}
\NoHyper
\footnote{\Letter: Corresponding authors.\\Emails: \texttt{yueyang22@mails.tsinghua.edu.cn}~~~  \texttt{gaohuang@tsinghua.edu.cn}~~~ \texttt{
doch@microsoft.com}
}
\endNoHyper
\addtocounter{footnote}{-1}
\endgroup

\maketitle

\begin{abstract}

Text and faces are among the most perceptually salient and practically important patterns in visual generation, yet they remain challenging for autoregressive generators built on discrete tokenization. A central bottleneck is the tokenizer: aggressive downsampling and quantization often discard the fine-grained structures needed to preserve readable glyphs and distinctive facial features.
We attribute this gap to standard discrete-tokenizer objectives being weakly aligned with text legibility and facial fidelity, as these objectives typically optimize generic reconstruction while compressing diverse content uniformly.
To address this, we propose \textit{InsightTok}, a simple yet effective discrete visual tokenization framework that enhances text and face fidelity through localized, content-aware perceptual losses. 
With a compact 16k codebook and a $16\times$ downsampling rate, InsightTok significantly outperforms prior tokenizers in text and face reconstruction without compromising general reconstruction quality. These gains consistently transfer to autoregressive image generation in \textit{InsightAR}, producing images with clearer text and more faithful facial details. Overall, our results highlight the potential of specialized supervision in tokenizer training for advancing discrete image generation.
\end{abstract}

\begin{figure}[h]
    \vspace{-5pt}
    \centering
    \includegraphics[width=1.0\linewidth]{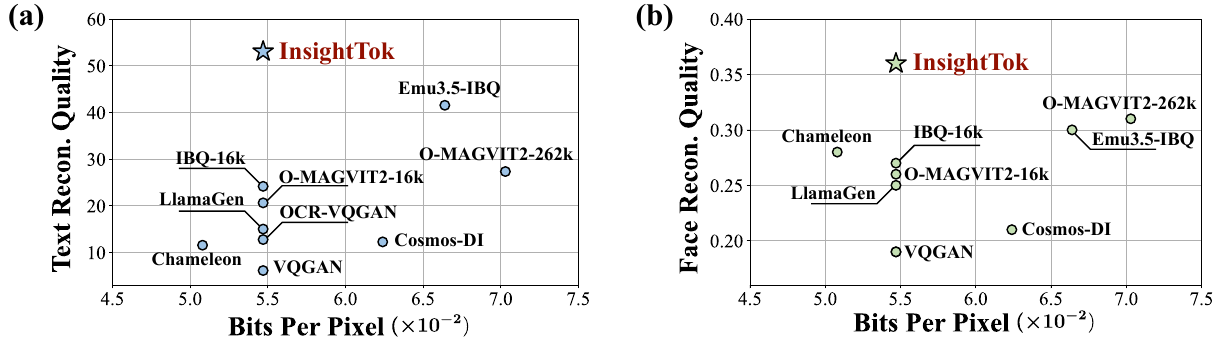}
    \vspace{-7pt}
    \caption{\textbf{Rate–distortion comparison of discrete tokenizers for text (a) and face (b) image reconstruction.} \textit{InsightTok} achieves state-of-the-art performance in text accuracy and face similarity, evaluated on TokBench~\cite{tokbench}. Compression rate is measured in bits per pixel.}
    \label{fig:teaser}
    \vspace{-10pt}
\end{figure}

\section{Introduction}

Discrete tokenization~\cite{vqvae} has become a cornerstone of autoregressive image generation~\cite{vqgan} and large-scale multimodal modeling~\cite{chameleon, emu3}, enabling unified processing of both visual and textual information. Central to this paradigm is the visual tokenizer, which maps continuous images into discrete token sequences. 
However, aggressive spatial downsampling and quantization often discard fine-grained details, making text and faces among the most prominent failure modes of existing tokenizers~\cite{tokbench}. This limitation is increasingly consequential as modern visual generative models are widely used in text- and face-centric scenarios such as graphic design, poster generation, and portrait synthesis. It is also perceptually salient: cognitive studies suggest that humans attend disproportionately to text and faces and are highly sensitive to distortions in these regions~\cite{wang2012attraction,cerf2009faces}.
% Improving text and face fidelity in discrete tokenization therefore essential for both practical applications and human-perceived visual quality.

Previous efforts typically address this issue by reducing compression, either by increasing the codebook size or the number of tokens per image~\cite{vqganlc, ibq, rqvae, unitok}, but these approaches incur substantial computational overhead and modeling complexity, and do not explicitly prioritize fidelity-critical structures. We argue that a key reason standard discrete tokenizers struggle with text and faces is \textit{insufficiently targeted supervision}. Common objectives such as pixel reconstruction loss and LPIPS~\cite{lpips} are designed for generic image reconstruction, but are poorly aligned with text readability and identity preservation. Moreover, text and face regions often occupy only a small fraction of an image, causing their training signals to be diluted by the surrounding scene. Consequently, conventional tokenizer training provides \textit{limited selective pressure} to preserve these high-value details under a tight discrete bottleneck.

% We argue that a key cause of poor text and face reconstruction is an \emph{importance misalignment} between standard tokenizer training objectives and human perceptual priorities. Cognitive studies suggest that humans attend disproportionately to text and faces and are highly sensitive to their distortions~\cite{wang2012attraction,cerf2009faces}. In contrast, conventional reconstruction and perceptual losses are applied uniformly across space and semantics. Since text and face instances often occupy a small fraction of the image, their supervision can be diluted by the rest of the scene, leaving the tokenizer with weak incentives to preserve these high-value details under a tight discrete bottleneck.

To address this gap, we propose \textit{InsightTok}, a simple and effective framework that explicitly enhances discrete visual representation learning of \textit{text} and \textit{faces}. InsightTok augments standard tokenizer training with localized, specialized perceptual losses for text and faces, computed on detected regions using domain-specific recognition models (Figure~\ref{fig:method}). These region-level objectives are combined with a weighted aggregation scheme (Section~\ref{sec:text}) that enables targeted improvements on perceptually critical content while maintaining general-purpose reconstruction. 

\begin{figure}[ht]
    \centering
    \includegraphics[width=0.9\linewidth]{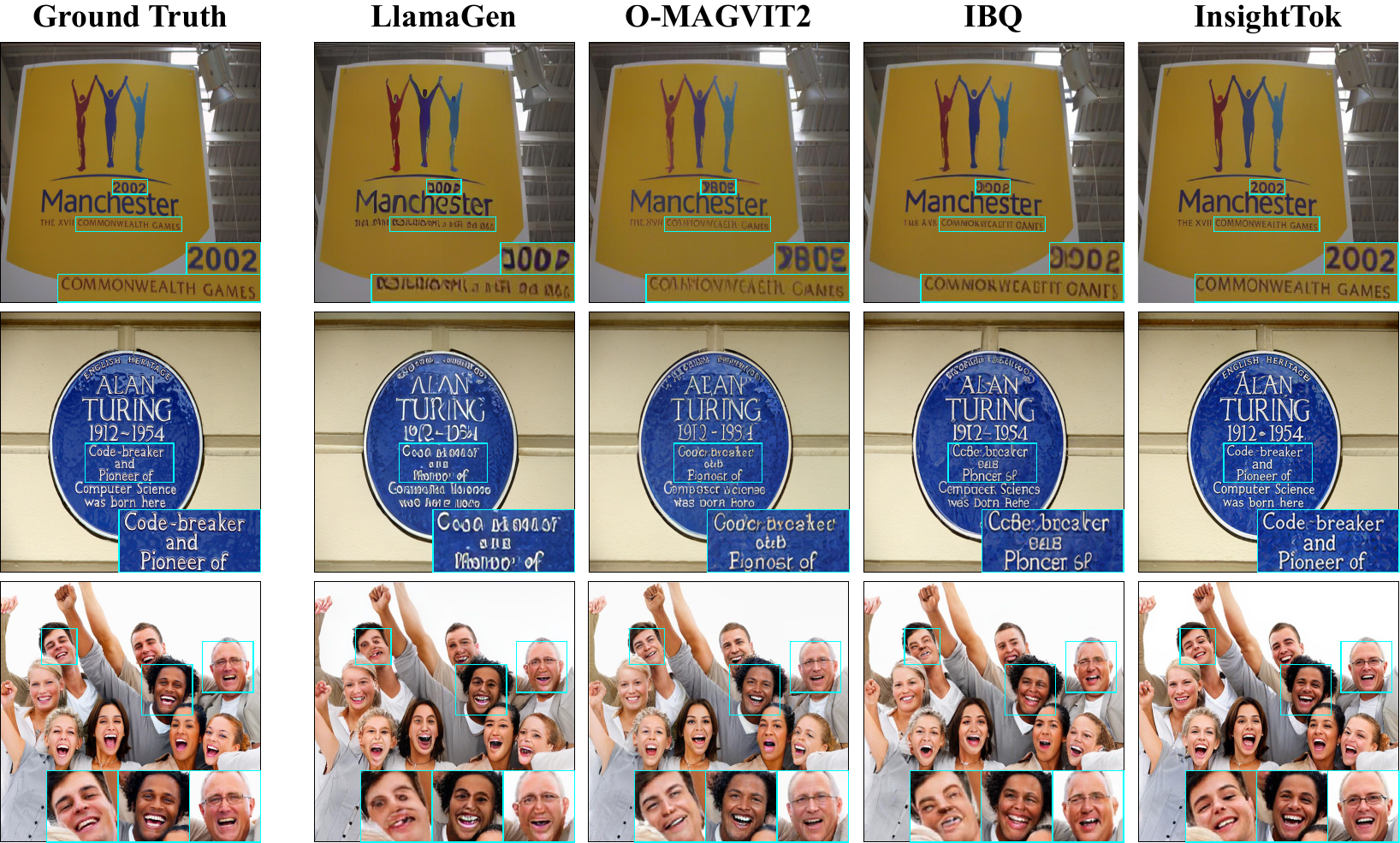}
    \vspace{-3pt}
    \caption{\textbf{Comparison of reconstruction quality} between InsightTok and existing tokenizers (LlamaGen~\cite{llamagen}, O-MAGVIT2~\cite{luo2024open} and IBQ~\cite{ibq}). All models use a codebook size of 16,384 and a downsampling rate of 16, evaluated at an image resolution of $512 \times 512$.}
    \vspace{-10pt}
    \label{fig:recon_vis}
\end{figure}

With a $16\times$ downsampling rate and a 16,384-entry codebook, InsightTok achieves substantial gains in text and face reconstruction (Figure~\ref{fig:teaser}, Figure~\ref{fig:recon_vis}), while remaining competitive on standard metrics (Table~\ref{tab:result_main}). We then develop \textit{InsightAR}, an autoregressive image generator trained on discrete codes produced by InsightTok, and show that the tokenizer improvements transfer consistently to text-to-image generation, producing images with clearer text and more faithful facial details.

Overall, our work offers a fresh perspective on tokenizer training by moving beyond the widely used VQGAN-style training supervision. It opens up a promising direction for incorporating richer, content-aware supervision into discrete representation learning.

\section{Related Work}

\textbf{Autoregressive image generation.} Autoregressive (AR) models generate images by factorizing the joint distribution over discrete visual tokens into a product of conditional next-token probabilities. With a discrete tokenizer~\cite{vqvae,vqgan} that converts an image into a low-resolution token grid, an AR Transformer is trained to model the sequence dependency, which has been scaled successfully for text-to-image synthesis~\cite{llamagen,emu3}. 
% Beyond strict left-to-right decoding, several \emph{generalized autoregressive} paradigms generate token maps via iterative refinement while predicting multiple tokens per step. For example, mask-based generative models~\cite{maskgit, muse} repeatedly fill in masked positions using bidirectional context, enabling parallel updates and faster sampling. Coarse-to-fine approaches like VAR~\cite{var} predicts token maps progressively across resolutions. 
The shared sequence modeling interface also aligns naturally with language modeling, enabling unified multimodal Transformers that jointly handle text and image tokens within a single architecture~\cite{chameleon,janus,emu35,luminamgpt}. 
% Across these families, generation quality is strongly bounded by the tokenizer: information lost at discrete tokenization time is difficult for any downstream model to recover, especially for semantically sensitive content such as text and faces.

\textbf{Discrete tokenizer designs.} VQ-VAE and its variants~\cite{vqvae,vqvae2} established the encoder--quantizer--decoder framework for learning discrete visual representations. VQGAN~\cite{vqgan} introduced the widely adopted training recipe that combines reconstruction, perceptual similarity~\cite{lpips}, and adversarial supervision to improve reconstruction fidelity. Subsequent work improved quantization and utilization in multiple directions, including codebook-free quantizers such as LFQ~\cite{magvitv2} and FSQ~\cite{fsq}, and methods that refine codebook learning and assignments~\cite{ibq,vqganlc,simvq}. Multi-code schemes reduce quantization error via residual quantization~\cite{rqvae}, and hierarchical generation strategies such as VAR~\cite{var} further leverage multi-scale token structures. Beyond reconstruction fidelity, several works incorporate higher-level semantics into tokenizers~\cite{tokenflow,lin2025toklip,zheng2025vision}.
% Beyond reconstruction fidelity, several works incorporate higher-level semantics into tokenizers, including coupling visual codes with semantic structure~\cite{tokenflow}, adding text--image alignment supervision~\cite{lin2025toklip,unitok}, or deriving tokenizers from frozen pretrained encoders~\cite{wang2024image,zheng2025vision}. 
Variable-length tokenization has also been explored to adapt token budgets to image content~\cite{titok,flextok}. Despite these advances, recent benchmarks suggest that discrete autoencoders still struggle to preserve fine-grained visual information~\cite{tokbench,vtbench}. \emph{In particular, text and faces remain persistent failure modes.} 
% Our work addresses this gap by introducing the first content-specialized tokenizer training framework that explicitly targets \emph{both} text readability and facial fidelity.

\textbf{Text and face generation.} Rendering legible text and faithful faces remains challenging in image synthesis, since small artifacts can harm readability and identity.
For text, recent methods predominantly target diffusion models, either by adding text-aware conditions (e.g., glyph/layout guidance) or applying OCR losses to emphasize correctness in text regions~\cite{tuo2023anytext,liu2024glyph,chen2023textdiffuser}.
For faces, prior work improves identity preservation by leveraging identity representations as supervision or conditioning in subject-driven generative models~\cite{shen2018faceid,wang2024instantid,li2024photomaker}.
Despite strong progress, these text- and face-specific techniques focus on diffusion models. Improving text and face quality in autoregressive generators is \emph{less explored}, as these models operate on discrete tokens rather than pixels. A related tokenizer-side approach, OCR-VQGAN~\cite{ocrvqgan}, introduces a global OCR-derived perceptual loss that has been primarily evaluated in diagram-oriented settings. 
Our approach instead improves the \emph{general-purpose} discrete tokenizer for autoregressive models via \emph{localized, content-specialized} perceptual supervision for \emph{both} text and faces, achieving significant gains while preserving overall reconstruction quality.

\section{Method}
\subsection{Preliminary: Discrete Image Tokenizer}

A discrete image tokenizer~\cite{vqvae} maps an image to a compact sequence of discrete symbols and reconstructs the image from these tokens. Operating in token space enables efficient autoregressive generative modeling. A tokenizer consists of an encoder $E$, a decoder $D$, and a vector quantizer $Q$ equipped with a learned codebook $C=\{\boldsymbol{e}_k\in \mathbb R^d\}_{k=1}^K$, where $K$ is the size of the codebook and $d$ is the dimension of codebook embeddings.  Given an input image $\bm x$, the encoder produces a downsampled latent map $\bm z = E(\bm x)$. The quantization layer $Q$ discretizes $\boldsymbol{z}$ by mapping each latent vector to its nearest entry in a learned codebook, producing a discrete token map $\hat{\boldsymbol{z}} = Q(\boldsymbol{z})$. The decoder then reconstructs the image as $\hat{\bm x}=D(\hat{\bm z})$.

\textbf{Training objectives.} The tokenizer components $(E,Q,C,D)$ are trained under a combination of complementary loss functions that balance pixel-level fidelity, effective codebook learning, and perceptual realism:
\begin{equation}
\label{eq:image-level}
\mathcal L_{\text{image}}
=\mathcal L_{\text{rec}}
+\beta\cdot \mathcal L_{\text{codebook}}
+\gamma\cdot \mathcal L_{\text{perc}}
+\eta\cdot \mathcal L_{\text{GAN}}.
\end{equation}
Here $\mathcal L_{\text{rec}}$ is an $\ell_1$ or $\ell_2$ reconstruction loss; $\mathcal L_{\text{codebook}}$ ensures effective codebook optimization; $\mathcal L_{\text{perc}}$ encourages similarity in a pretrained feature space to preserve semantic and textural structure; and $\mathcal L_{\text{GAN}}$ employs an auxiliary discriminator to reduce artifacts and improve visual fidelity. Scalars $\beta,\gamma,\eta$ control the relative contributions of each term.

% The image tokenizer, consisting of the encoder $E$, quantization layer $Q$, and decoder $D$, is trained under a combination of complementary loss functions that jointly encourage faithful reconstruction, perceptual quality, and effective discrete representation learning. The commonly used image-level tokenizer training objective $\mathcal L_{\text{image}}$ is defined as:
% \begin{equation}
% \label{eq:image-level}
% \mathcal L_{\text{image}} = \mathcal L_{\text{rec}} 
% + \beta \cdot \mathcal L_{\text{codebook}} 
% + \gamma \cdot \mathcal L_{\text{perc}} 
% + \eta \cdot \mathcal L_{\text{GAN}},
% \end{equation}
% where $\mathcal L_{\text{rec}}$ is the reconstruction loss that enforces pixel-level consistency between the input image $\boldsymbol{x}$ and its reconstruction $\hat{\boldsymbol{x}}$, which typically implemented as an $\ell_1$ or $\ell_2$ loss; $\mathcal L_{\text{codebook}}$ denotes codebook optimization loss; perceptual loss $\mathcal L_{\text{perc}}$ encourages perceptual similarity and learn fine-grained structure of the image; adversarial regularization $\mathcal L_{\text{GAN}}$ prevent artifacts and increase image fidelity with a jointly trained discriminator; $\beta$, $\gamma$, and $\eta$ are weighting coefficients that balance the contributions of different loss terms.

\textbf{Codebook optimization.} We follow the standard \emph{vector quantization (VQ)} formulation and update the codebook embeddings using an \emph{exponential moving average (EMA)} scheme, which has been shown to be stable and effective for large codebooks and large embedding dimensions. To couple the encoder outputs to their assigned codewords, we use the standard commitment loss $\mathcal L_{\text{codebook}}=\big\|\bm z-\operatorname{sg}(\hat{\bm z})\big\|_2^2$,
% \begin{equation}
% \mathcal L_{\text{codebook}}=\big\|\bm z-\operatorname{sg}(\hat{\bm z})\big\|_2^2,
% \end{equation}
where $\operatorname{sg}(\cdot)$ denotes the stop-gradient operator.
% and pass gradients to the encoder through the discrete assignments via the straight-through estimator. 
To improve codebook utilization, we adopt a \textit{restart strategy} that periodically reinitializes codewords that remain unused for extended periods. Full details are provided in Appendix~\ref{app:quant}.

% \textbf{Perceptual Loss.} The perceptual loss $\mathcal L_{\text{perc}}$ measures the discrepancy between $\boldsymbol{x}$ and $\hat{\boldsymbol{x}}$ in a high-level feature space extracted by a pretrained network (e.g., VGG). Unlike pixel-wise losses, this term emphasizes semantic consistency and perceptual similarity. In the tokenizer community, LPIPS~\cite{lpips} is the most widely used perceptual loss, which is a perceptual similarity estimator learned from human annotations, which takes the form: 
% \begin{equation}
% \operatorname{\mathcal L_{\text{perc}}} = \sum_l \frac 1 {h_lw_l}\big\|\bm w_l \odot \left(F^{(l)}(\hat{\bm x}) - F^{(l)}(\bm x)\right)\big\|
% \end{equation}
% where $F^{(l)}$ is the $l$-th layer intemediate feature of a VGG network and $w_l$ is the learned coefficient.

% \textbf{Adversarial Loss.} To further enhance visual realism and suppress reconstruction artifacts, an adversarial loss $\mathcal L_{\text{GAN}}$ is adopted, as first introduced in VQ-GAN~\cite{vqgan}. A discriminator is trained to distinguish real images from reconstructed ones, while the tokenizer is optimized to generate reconstructions that are indistinguishable from real images, leading to sharper and more realistic outputs.

\textbf{Perceptual loss.}
Pixel-level reconstruction losses alone often yield overly smooth outputs, so visual tokenizers commonly incorporate perceptual supervision that compares $\bm x$ and $\hat{\bm x}$ in a pretrained feature space, such as LPIPS~\cite{lpips}:
\begin{equation}
\mathcal L_{\text{perc}}
=\sum_{l}\frac{1}{H_l W_l}\left\|
\bm w_l\odot\big(F^{(l)}(\hat{\bm x})-F^{(l)}(\bm x)\big)
\right\|^2,
\end{equation}
where $F^{(l)}(\cdot)$ are deep features of a pretrained VGG network, $(H_l,W_l)$ is its spatial resolution, and $\bm w_l$ are learned channel-wise weights. While effective at improving overall perceptual quality, this loss is derived from a patch-similarity dataset~\cite{lpips} that does not fully capture glyph readability or facial features. Moreover, by averaging errors across the entire image, it can underemphasize text and face regions that are small yet perceptually critical.

% We address this with \textit{localized} perceptual losses targeting text readability and facial fidelity, complementing the general-purpose LPIPS objective.

% This loss improves perceptual fidelity by emphasizing high-level structure and texture over exact pixel alignment. A limitation of a single \emph{global} perceptual loss is that it averages errors over the entire image and can underweight small but semantically critical regions, notably \emph{text} and \emph{faces}. Motivated by this, we extend the standard perceptual objective with \emph{specialized} perceptual losses that explicitly target text readability and facial fidelity, beyond the general-purpose LPIPS.

%\subsection{InsightTok: Preserve Text and Face Under Aggressive Compression}

\begin{figure}
\centering
\includegraphics[width=1.0\linewidth]{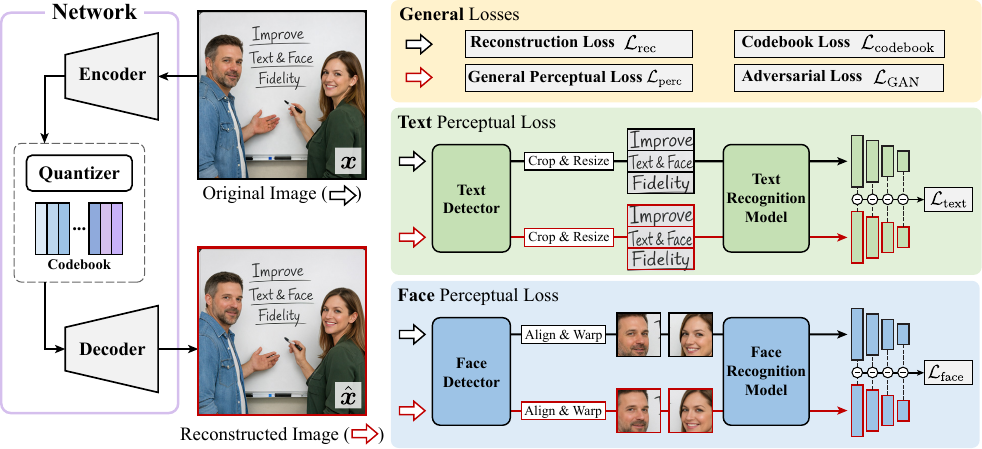}
\vspace{-5pt}
    \caption{\textbf{Illustration of the proposed framework.}
    % Discrete tokenizers are trained with a combination of losses including $\mathcal L_{\text{rec}}$, $\mathcal L_{\text{perc}}$, $\mathcal L_{\text{codebook}}$, and $\mathcal L_{\text{GAN}}$.
    In addition to standard tokenizer losses, \textit{InsightTok} introduces localized, content-aware perceptual losses, $\mathcal L_{\text{text}}$ and $\mathcal L_{\text{face}}$, to prioritize critical text and face regions. These regions are detected and sampled from both the original and reconstructed images, and processed through domain-specific recognition models to compute the perceptual losses.}
    \label{fig:method}
    \vspace{-10pt}
\end{figure}

\subsection{InsightTok}

% \noindent\textbf{Objective.} 
% % Discrete visual tokenizers typically suffer from information loss due to discretization in a spatially lower-dimensional representation. Standard training objectives (pixel reconstruction, generic perceptual similarity, and adversarial losses) treat all visual signals as equally informative and ignore semantic distinctions within visual content.
% Discrete visual tokenizers operate under a tight representational bottleneck, where uniform reconstruction objectives fail to prioritize perceptually important content.
% In this section, we argue that the representational capacity of visual tokenizers remains underexploited when optimized solely with such general-purpose objectives. To address this limitation,

\textbf{Overview.} Standard tokenizer training objectives typically treat diverse semantic content uniformly, and are often insufficiently sensitive to subtle differences in text readability and facial similarity. To address this limitation, the InsightTok framework augments conventional tokenizer training with targeted, content-aware supervision. As illustrated in Figure~\ref{fig:method}, InsightTok adds two content-aware perceptual terms: a text perceptual loss $\mathcal L_{\text{text}}$ (Section~\ref{sec:text}) and a face perceptual loss $\mathcal L_{\text{face}}$ (Section~\ref{sec:face}). These terms complement the conventional image-level tokenizer objective $\mathcal L_{\text{image}}$, as defined in Eq.~\ref{eq:image-level}. The overall optimization objective is given by:
\begin{equation}
\label{eq:loss}
\mathcal L_{\text{InsightTok}} = \mathcal L_{\text{image}} + \alpha_1\cdot \mathcal L_{\text{text}} + \alpha_2\cdot \mathcal L_{\text{face}},
\end{equation}
where $\alpha_1$ and $\alpha_2$ are scalar loss weights.

\subsubsection{Text Perceptual Loss}
\label{sec:text}
% \textbf{Text Detection.} Our effort to improve text reconstruction start with collecting text-rich images from LAION~\cite{laion} and annotate textual regions in all the images with a text detector~\cite{db}. In the resulting dataset, each training image $\boldsymbol{x}$ is associated with a set of text bounding boxes, denoted as $\{\boldsymbol{b}_{\text{text}}^{n}\}_{n=1}^{N}$, where $N$ is the number of detected regions. 
\textbf{Text detection.} We first curate text-rich training images from LAION~\cite{laion}. For each image $\bm x$, we detect text instances using a text detector~\cite{db}, producing a set of bounding boxes $\{\bm b_{\text{text}}^{n}\}_{n=1}^{N}$, where $N$ denotes the number of detected text regions.

% \textbf{Text Region Extraction.} The training image $\boldsymbol{x}$ is fed into the tokenizer to obtain a reconstructed image $\hat{\boldsymbol{x}}$ of the same resolution. We crop the corresponding text regions corresponding to the text bounding boxes $\boldsymbol{b}_{\text{text}}^n$ from both $\boldsymbol{x}$ and $\hat{\boldsymbol{x}}$, respectively, resulting in two sets of text regions: $\{\boldsymbol{r}_{\text{text}}^n\}_{n=1}^{N}$ from the ground-truth image, and $\{\hat{\boldsymbol{r}}_{\text{text}}^n\}_{n=1}^{N}$ from the reconstructed image.  

\textbf{Text region extraction.} Given a training image $\bm x$, the tokenizer produces a reconstruction $\hat{\bm x}$ at the same resolution. We crop corresponding regions from $\bm x$ and $\hat{\bm x}$ using each box $\bm b_{\text{text}}^n$, yielding paired patches $\{ \bm r_{\text{text}}^n \}_{n=1}^{N}$ and $\{ \hat{\bm r}_{\text{text}}^n \}_{n=1}^{N}$.

\textbf{Text-aware supervision.} To measure reconstruction quality specifically for text, we compare each pair $(\bm r_{\text{text}}^n,\hat{\bm r}_{\text{text}}^n)$ in the feature space of a pretrained text recognition network~\cite{fang2021read}, denoted $F_{\text{text}}(\cdot)$. Each crop is resized to a canonical banner resolution of $32\times128$ before being fed into $F_{\text{text}}(\cdot)$. We extract intermediate features from $L$ hidden layers, denoted $\{F_{\text{text}}^{(l)}(\cdot)\}_{l=1}^L$ and use $L=5$ by default. We define the region-level text perceptual loss as
\begin{equation}
\label{eq:text-loss-region}
\mathcal L^{n}_{\text{text}} =
\frac{1}{L}\sum_{l=1}^{L}\frac{1}{H_l W_l}
\big\|F_{\text{text}}^{(l)}(\bm r_{\text{text}}^n)-F_{\text{text}}^{(l)}(\hat{\bm r}_{\text{text}}^n)\big\|^2,
\end{equation}
where $(H_l,W_l)$ is the spatial size of the $l$-th feature map.

\textbf{Aggregation across regions.} We aggregate region losses as
\begin{equation}
\label{eq:text-loss}
\mathcal L_{\text{text}}=\sum_{n=1}^{N} w^n_{\text{text}}\cdot\mathcal L^{n}_{\text{text}},
\end{equation}
where weights $w^n_{\text{text}}$ control the contribution of each text region. Specifically, we define the weights to be proportional to the region size, namely
\begin{equation}
    w^n_{\text{text}}=\mathrm{Area}(\bm b_{\text{text}}^n)/\mathrm{Area}(\bm x),
\end{equation}
where $\bm x$ is the original image and $\text{Area}(\cdot)$ computes the area of the bounding box/image.  This area-based weighting prevents tiny text instances from dominating the overall objective:\textit{ small crops are inherently harder to reconstruct under discrete tokenization and often yield disproportionately large feature discrepancies}. Down-weighting them stabilizes training and balances contributions across text regions of different scales.

\subsubsection{Face Perceptual Loss}
\label{sec:face}

\textbf{Face and landmark localization.} Similar to text, we first perform face detection on the LAION dataset~\cite{laion} and retain only images in which at least one face is successfully detected. For each training image $\bm x$, the face detector~\cite{deng2019arcface} outputs a set of detected face instances,
$\big\{(\bm b_{\text{face}}^m,\{\bm p_k^m\}_{k=1}^{5})\big\}_{m=1}^{M}$,
where $M$ is the number of detected faces in the image. Here,  $\bm b_{\text{face}}^m$ denotes the face bounding box and $\{\bm p_k^m\}_{k=1}^{5}$ are the associated five facial landmarks (left/right eye, nose, and two mouth corners). These landmarks provide a reliable geometric reference for subsequent face alignment.

% All detected faces are denoted as $\{(\boldsymbol{b}_{\text{face}}^m, \{\boldsymbol{p}^m_k\}_{k=1}^{5})\}_{m=1}^{M}$, where $M$ is the total number of detected faces.

% \begin{figure}[t]
%     \centering
%     \includegraphics[width=0.7\linewidth]{figs/warp_v3.pdf}
%     \vspace{-5pt}
% \caption{\textbf{Illustration of face alignment.} The facial region is warped to align with the canonical template based on optimal landmark matching.}
%     \label{fig:warp}
% \end{figure}

\begin{wrapfigure}{r}{0.6\linewidth}
   \centering
   \vspace{-12pt}
    \includegraphics[width=1.0\linewidth]{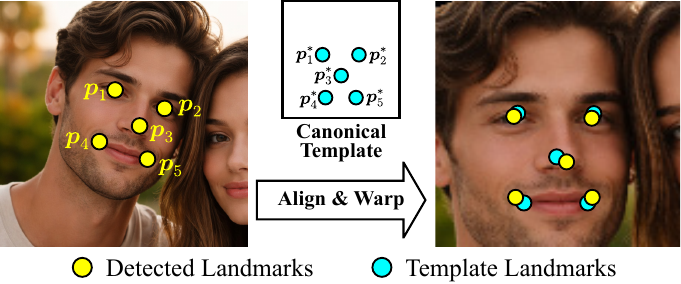}
    \vspace{-12pt}
\caption{\textbf{Illustration of face alignment.} The facial region is warped to align with the canonical template based on optimal landmark matching.}
\label{fig:warp}
\end{wrapfigure}

% \textbf{Face Region Extraction.} Given a training image $\boldsymbol{x}$ and its reconstruction $\hat{\boldsymbol{x}}$, we crop face regions from both images according to each face bounding box $\boldsymbol{b}_{\text{face}}^m$, producing the ground-truth face regions $\{\boldsymbol{r}_{\text{face}}^m\}_{m=1}^{M}$ and the reconstructed face regions $\{\hat{\boldsymbol{r}}_{\text{face}}^m\}_{m=1}^{M}$.

% [TODO!!!!] There's no extra croping operation before the alignment!

\textbf{Face alignment and region extraction.} To reduce variations in pose, scale, and in-plane rotation, we align each detected face to a canonical template (Figure~\ref{fig:warp}). Given detected landmarks $\{\bm p_k\}_{k=1}^{5}$ and template landmarks $\{\bm p_k^{\ast}\}_{k=1}^{5}$, we estimate a similarity transform $\mathcal T(\bm u)= s\bm R\bm u+\bm t$, where $\bm u\in\mathbb{R}^2$ is an image coordinate, $s$ is a scalar scale, $\bm R\in\mathbb{R}^{2\times 2}$ is a rotation matrix, and $\bm t\in\mathbb{R}^2$ is a translation.
The transformation parameters are obtained by minimizing the landmark alignment error:
\begin{equation}
\min_{s,\bm R,\bm t}\ \sum_{k=1}^{5}\left\|s\bm R\bm p_k+\bm t-\bm p_k^{\ast}\right\|^2.
\end{equation}
We then extract aligned face patches from both the input image $\bm x$ and its reconstruction $\hat{\bm x}$ via inverse warping into the canonical coordinate system:
\begin{equation}
\bm r_{\text{face}}[\bm c]=\bm x\left[\mathcal T^{-1}(\bm c)\right],\quad\hat{\bm r}_{\text{face}}[\bm c]=\hat{\bm x}\left[\mathcal T^{-1}(\bm c)\right],
\end{equation}
where $\bm c$ indexes the canonical face canvas (typically $112^2$) and $[\cdot]$ denotes pixel sampling. Here $\bm r_{\text{face}}$ and $\hat{\bm r}_{\text{face}}$ are the aligned face regions extracted from $\bm x$ and $\hat{\bm x}$, respectively.

\textbf{Face supervision.} We measure face-specific fidelity using a face recognition network. Concretely, we adopt the ResNet50-based face recognition model~\cite{deng2019arcface}, denoted $F_{\text{face}}(\cdot)$, and extract $L$ intermediate feature maps from each aligned pair $(\bm r_{\text{face}}^m,\hat{\bm r}_{\text{face}}^m)$.
The face perceptual loss $\mathcal{L}_{\text{face}}$ is defined as:
\begin{equation}
\label{eq:face-loss}
\begin{aligned}
\mathcal{L}_{\text{face}}^m
&=
\frac{1}{L}
\sum_{l=1}^{L}
\frac{1}{H_l W_l}
\big\|
F_{\text{face}}^{(l)}(\boldsymbol{r}_{\text{face}}^m)
-
F_{\text{face}}^{(l)}(\hat{\boldsymbol{r}}_{\text{face}}^m)
\big\|^2,\\
\mathcal{L}_{\text{face}} &= \sum_{m=1}^M w^m_{\text{face}}\cdot \mathcal{L}_{\text{face}}^m,
\end{aligned}
\end{equation}
where $(H_l, W_l)$ denotes the spatial resolution of the $l$-th feature map
$F_{\text{face}}^{(l)}(\boldsymbol{r}_{\text{face}}^m)$ (or
$F_{\text{face}}^{(l)}(\hat{\boldsymbol{r}}_{\text{face}}^m)$),
and weights $w^m_{\text{face}}$ control the contribution of each face instance.
We set $w^m_{\text{face}}=\text{Area}(\bm b_{\text{face}}^m)/\text{Area}(\bm x)$, consistent with Section~\ref{sec:text}, to balance faces of different scales and prevent small, difficult cases from dominating the overall objective.

\subsection{InsightAR}
We adopt a standard autoregressive (AR) image modeling pipeline to model the discrete tokens produced by InsightTok for text-to-image generation. 
Given an image $\bm x$, InsightTok encodes it into a $16\times$ downsampled token grid and rasterizes the grid into a sequence $\bm t=({t_1,\ldots,t_n})$, where each token $t_i$ indexes the tokenizer vocabulary. Conditioned on an input text prompt $T$, InsightAR parameterizes the joint distribution over image tokens with a Transformer~\cite{vaswani2017attention} and trains via next-token prediction:
\begin{equation}
p(\bm t \mid T)=\prod_{i=1}^{n} p(t_i \mid t_{<i}, T).
\end{equation}
At generation time, we sample tokens sequentially from $p(t_i \mid t_{<i}, T)$ and decode the completed token map back to an image using the InsightTok decoder.
The architecture of InsightAR largely follows Janus-Pro~\cite{janus}, except that the tokenizer is replaced with InsightTok in order to improve text and face fidelity.

% \begin{equation}
% t_n \sim p(t_n \mid t_1, \dots, t_{n-1}; T),
% \end{equation}

\section{Implementation}
\textbf{InsightTok} follows the convolutional architecture of VQGAN~\cite{vqgan} with a downsampling rate of $16$. The model contains 426M parameters. The codebook size is set to 16,384, with each embedding having a dimensionality of 256. Training proceeds in three stages. First, the tokenizer is pretrained for 200k steps using standard objectives, including reconstruction, general perceptual, and adversarial losses. The tokenizer is then further trained for 40k steps with the proposed text and face perceptual losses, $\mathcal L_{\text{text}}$ and $\mathcal L_{\text{face}}$, on curated subsets of LAION~\cite{chen2023textdiffuser, zheng2022general}. Finally, the encoder and quantizer are frozen and the decoder is fine-tuned for an additional 40k steps to refine reconstruction quality.
% The model is trained with a $256^2$ image resolution, a batch size of 512, and a learning rate of $1 \times 10^{-4}$. 
Additional implementation details are provided in Appendix~\ref{app:impl_tok}.

\textbf{InsightAR} is trained on discrete token sequences produced by InsightTok. The architecture and training procedure follow the Janus-Pro~\cite{janus} framework. An MLP adapter connects the visual tokenizer to a multimodal large language model with 7B parameters. 
% The training occurs in two stages: first, the adapter and visual token prediction head are warmed up for 40k steps with the transformer frozen; second, the full model is trained for one epoch on large-scale text-to-image data. 
The training set is a filtered mixture of LAION~\cite{laion}, Flux-Reason-6M~\cite{fang2025flux}, Echo-4o~\cite{ye2025echo}, and synthetic text rendering data\footnote{\href{https://github.com/GbotHQ/ocr-dataset-rendering}{https://github.com/GbotHQ/ocr-dataset-rendering}}, totaling around 150M images. All images are transformed to $512^2$ resolution and represented by 1,024 tokens.
For comparison, we also train an autoregressive model using the LlamaGen tokenizer~\cite{llamagen}, which is used in the original Janus-Pro, under the same setup, denoted as \textit{LlamaGenTok-AR}. Additional training details are provided in Appendix~\ref{app:impl_ar}.

\begin{table*}[t]
    \centering
    \caption{\textbf{Reconstruction performance of InsightTok and existing discrete visual tokenizers.} Compression rate is measured in bits per pixel (BPP), defined as the number of bits used to represent an image divided by its spatial resolution. Text and face reconstruction are evaluated on TokBench~\cite{tokbench}, using text accuracy (T-ACC), text normalized edit distance (T-NED), and face similarity (F-Sim), following the official protocols.
    Subscripts `s' and `m' (mean) denote metrics averaged over small-scale and all instances, respectively. General reconstruction performance is evaluated on the ImageNet~\cite{imagenet} validation set, with rFID and PSNR reported. All images are at $512 \times 512$ resolution. \textbf{Bold} and \underline{underlined} denote the best and second-best results, respectively.
    }
    \vspace{-3pt}
    \label{tab:result_main}
    \resizebox{1.0\linewidth}{!}{
    \centering
    \setlength{\tabcolsep}{2pt}
    \begin{tabular}{l|ccc|cccc|cc|cc}
    \toprule
       \multirow{2}{*}{\textbf{Method}} & \multirow{2}{*}{\textbf{\#Tokens}} & \multirow{2}{*}{\textbf{Codebook}}  & \multirow{2}{*}{\textbf{BPP}} & \multicolumn{4}{|c|}{\textbf{Text (\%) $\uparrow$}} & \multicolumn{2}{|c|}{\textbf{Face $\uparrow$}} &  \multicolumn{2}{|c}{\textbf{General}}\\
       % \cmidrule(lr){6-7} \cmidrule(lr){8-9} \cmidrule(lr){10-11}
       &  & & & T-ACC$_s$ & T-ACC$_m$ &  T-NED$_s$ & T-NED$_m$ & F-Sim$_s$ & F-Sim$_m$  & rFID$\downarrow$ & PSNR$\uparrow$ \\
        % \midrule
        % \multicolumn{12}{c}{\textit{Resolution: 512 }$\times$\textit{ 512}} \\
        \midrule
        Chameleon~\cite{chameleon}  & 1,024& 8,192& 0.0508&0.60 & 11.55& 7.63&  26.80&  0.13& 0.28& 0.93& 22.53\\
        VAR~\cite{var} & 2,240 & 4,096& 0.1025 &3.71& 29.31& 18.01& 50.00&  0.14&  0.32& 0.58 & 23.43\\
        % TokenFlow~\cite{tokenflow} & MSRQ & & $32768$& & 1.06 &  17.40& 10.00&35.49 & 0.11& 0.24& &\\
        % VILA-U & RQ & & & & & & & & &\\
        % Infinity~\cite{infinity}& BSQ & & $2^32$& & & & & & & & &\\
        Cosmos-DI~\cite{agarwal2025cosmos} & 1,024& 64,000 & 0.0624 & 0.52& 12.26& 7.61& 27.95& 0.09& 0.21& 1.36 & 22.35 \\
        O-MAGVIT2-262k~\cite{luo2024open} & 1,024 & 262,144  & 0.0703 &3.02 &  27.33&16.87 & 47.28 &0.13&0.31& 0.65 & 23.98 \\
        Emu3.5-IBQ~\cite{emu35}& 1,024 & 131,072  & 0.0664 & 12.99 & 41.52 & 39.92 & 65.39 & 0.14& 0.30 & 0.46 & 23.82 \\ 
        \midrule
        VQGAN~\cite{vqgan} &\multirow{6}{*}{1,024} & \multirow{6}{*}{16,384} & \multirow{6}{*}{0.0547} & 0.15& 6.12 &  5.20& 17.32&  0.08& 0.19& 1.27 & 22.24\\
        OCR-VQGAN~\cite{ocrvqgan} & &  &  & 1.13& 12.76& 10.24& 28.58& 0.08& 0.19& 16.67& 21.49\\
        LlamaGen-T2I~\cite{llamagen} && & & 0.67& 15.01&  7.76&  30.44& 0.11& 0.25 & \underline{0.69} & 22.80 \\
        O-MAGVIT2-16k~\cite{luo2024open} & &  &  & 1.62 & 20.62& 12.71& 39.96& 0.11& 0.26& 1.20 & 23.31 \\
        IBQ-16k~\cite{ibq}& &   &  & \underline{2.28} & \underline{24.16} & \underline{14.75} & \underline{43.66} & \underline{0.12}& \underline{0.27}& \textbf{0.64} & \underline{23.39} \\ 
        \methodname{}& &  & & \textbf{16.44}& \textbf{53.05}& \textbf{42.31}& \textbf{71.40}& \textbf{0.16}& \textbf{0.36} & \underline{0.69}& \textbf{23.64}\\
        % \midrule
        % {\color{gray}DC-AE~\cite{dcae}} & {\color{gray}256} & {\color{gray}32 chans.} & {\color{gray}1.0000} &  {\color{gray}5.31} & {\color{gray}38.25}&{\color{gray}20.91} &{\color{gray}56.22} & {\color{gray}0.16} &  {\color{gray}0.39}& {\color{gray}0.34}& {\color{gray}25.81} \\
        % {\color{gray}VA-VAE~\cite{vavae}} & {\color{gray}1024} & {\color{gray}32 chans.}  & {\color{gray}4.0000} &  {\color{gray}12.72}& {\color{gray}53.30}
        % &{\color{gray}34.80} & {\color{gray}69.65}& {\color{gray}0.31}& {\color{gray}0.57}& {\color{gray}0.27}& {\color{gray}28.02} \\
        % {\color{gray}SD-XL~\cite{sdxl}} & {\color{gray}4096} & {\color{gray}4 chans.}  & {\color{gray}2.0000} & {\color{gray}16.53} & {\color{gray}56.86}& {\color{gray}40.43}& {\color{gray}72.55} &  {\color{gray}0.29}& {\color{gray}0.55}& {\color{gray}0.28}& {\color{gray}28.34}\\
        \bottomrule
      \end{tabular}}
      \vspace{-5pt}
\end{table*}

\begin{figure}[t]
    \centering

    \includegraphics[width=1.0\linewidth]{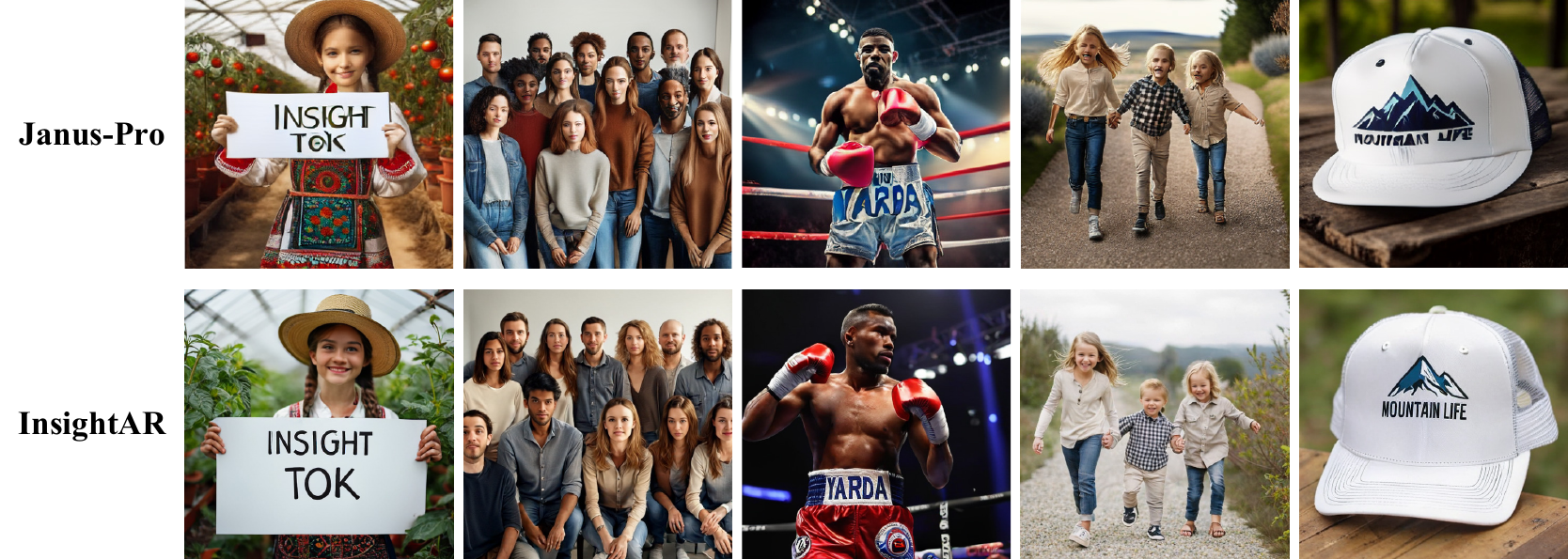}
    \vspace{-10pt}
    \caption{\textbf{Comparison of images generated by Janus-Pro and InsightAR.} Appendix~\ref{app:vis} provides more visualizations.}
    \label{fig:gen_vis}
    \vspace{-10pt}
\end{figure}

\section{Experiments}

\subsection{Image Reconstruction}
\textbf{Evaluation protocols.} We evaluate text and face reconstruction using the TokBench~\cite{tokbench} benchmark, which defines challenging in-the-wild reconstruction tasks for textual content and human faces. Text reconstruction is assessed with an OCR toolbox~\cite{trullemans2016doctr}, using text accuracy (T-ACC) and normalized edit distance (T-NED) against ground truth annotations. Face reconstruction quality is measured by the similarity score~\cite{deng2019arcface} between reconstructed faces and their corresponding ground truth. In addition, general reconstruction performance is evaluated on the ImageNet validation set using rFID and PSNR. Most of the baseline tokenizers considered are designed for text-to-image generation and are trained on diverse image corpora \textit{beyond ImageNet}. Full details are presented in Appendix~\ref{app:recon}.

\textbf{Results.} As shown in Table~\ref{tab:result_main}, InsightTok outperforms existing discrete tokenizers across all the text and face reconstruction metrics.
% In particular, InsightTok achieves a 28\% improvement in text accuracy (T-ACC) and a 0.09 increase in face similarity compared to the second-best method, IBQ~\cite{ibq}, at the same compression ratio. Our model also consistently outperforms Emu3.5-IBQ~\cite{emu35}, which uses a significantly larger codebook (131k entries). 
At the same compression ratio, InsightTok improves text accuracy (T-ACC) by 28.89 percentage points and face similarity by 0.09 over the second-best method, IBQ~\cite{ibq}. Our model also consistently outperforms Emu3.5-IBQ~\cite{emu35} despite its much larger codebook of 131k entries.
Moreover, InsightTok achieves competitive results on general metrics, reaching a PSNR of over 23.6, demonstrating that our method does not sacrifice the quality of non-textual and non-facial regions. These advantages are further highlighted in Figure~\ref{fig:recon_vis}, where the visual comparisons of InsightTok’s reconstructions against the baselines clearly show superior quality.

\subsection{Autoregressive Text-to-Image Generation}

\begin{table*}[t]
    \centering
    %\captionsetup{font={footnotesize}}
    \caption{\textbf{Image generation performance of InsightAR and existing autoregressive models.} Face generation is evaluated by generating a crowd of up to twenty individuals, with quality measured by the MagFace embedding norm~\cite{meng2021magface}. Text generation is assessed by rendering a paragraph of up to 300 characters on blank backgrounds, with normalized edit distance (NED) as the metric. General text-to-image ability is evaluated on GenEval~\cite{geneval} and DPG-Bench~\cite{dpg}.}
    \label{tab:main_gen}
    \vspace{-3pt}
    \resizebox{1.0\linewidth}{!}{
    \setlength{\tabcolsep}{2.5pt}
    \begin{tabular}{lc|cc|c|c|cc}
    \toprule
    \multirow{2}{*}{\textbf{Model}} & \multirow{2}{*}{\textbf{\#Params}} & \multirow{2}{*}{\textbf{Resolution}} & \multirow{2}{*}{\textbf{\#Tokens}} & \textbf{Face} & \textbf{Text} & \multicolumn{2}{|c}{\textbf{General}}  \\
    & & & & MagFace-Score$\uparrow$ & NED (\%)$\uparrow$ & GenEval$\uparrow$ & DPG-Bench$\uparrow$ \\
    \midrule
    \multicolumn{2}{l|}{\textit{$8\times$ Downsample Rate, Larger Resolution}} & & & & & & \\
    % Chameleon \cite{chameleon} & 7B & 0.39 & - \\
    % TokenFlow-XL \cite{tokenflow} & TokenFlow & 14B   & 0.63 & 73.38 & & &\\
    Emu3 \cite{emu3} & 8B  & 720 & 8,100 & 23.70& 17.82 & 0.66 & 80.60\\
    Lumina-mGPT2.0 \cite{luminamgpt}  & 7B & 768 & 9,216  & 23.09& 27.53 & 0.80 & 84.30\\
    \midrule
    % \multicolumn{9}{l}{\textit{$\times16$ Downsample Rate}} \\
    \multicolumn{2}{l|}{\textit{$16\times$ Downsample Rate}} &  & & & & \\
    LlamaGen \cite{llamagen} & 0.8B & 512 & 1,024  &22.37& 18.78 & 0.32 & 65.16\\
    Show-o \cite{showo} & 1.3B  & 512& 1,024& \underline{22.61}& 22.45 & 0.68 & 67.27\\
    Janus-Pro \cite{janus} & 7B & 384 & 576  & 22.09& 32.29 & 0.80 & \textbf{84.19}\\
    LlamaGenTok-AR & 7B & 512 & 1,024 & 22.29 & \underline{79.86} & \underline{0.81} & 83.78 \\
    % Emu3.5 \cite{emu35} & 34B & 0.86 & 88.26 \\
    % \midrule
    InsightAR & 7B & 512 & 1,024 & \textbf{23.33}& \textbf{95.83}& \textbf{0.82} & \underline{84.11} \\
        \bottomrule
    \end{tabular}
    }
    \vspace{-7pt}
\end{table*}

\begin{figure}[t]
    \centering
    \includegraphics[width=0.9\linewidth]{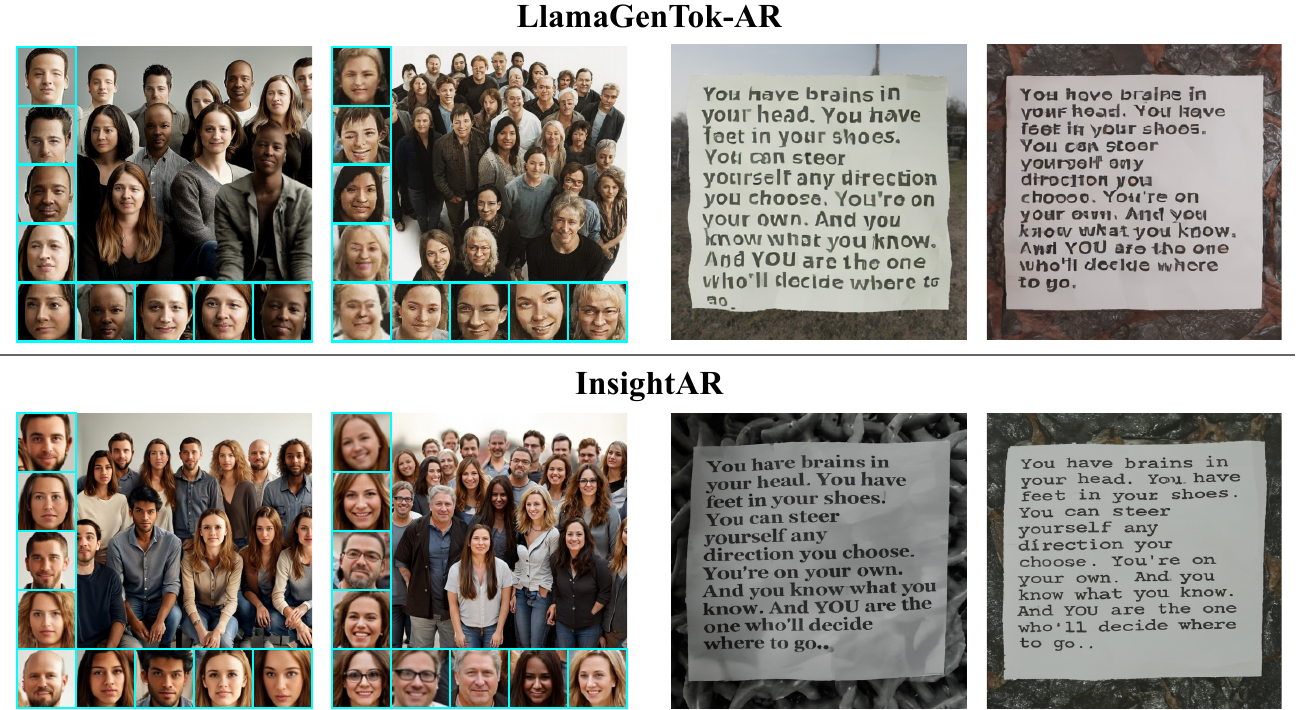}
    \vspace{-3pt}
    \caption{Comparison of face quality (left) and long text rendering (right) between images generated by \textit{LlamaGenTok-AR} and \textit{InsightAR}. 
    % The results highlight that the tokenizer's facial reconstruction capability significantly impacts the quality of face generation.
    }
    \label{fig:face_text_gen}
    \vspace{-10pt}
\end{figure}

% In this setting, each individual's face appears smaller, which emphasizes the tokenizer's ability to effectively represent facial details.
\textbf{Face generation.} We evaluate face generation quality in a challenging crowd-generation setting, where models synthesize images containing many individuals (examples shown in the left part of Figure~\ref{fig:face_text_gen} and Appendix~\ref{app:gen_face}). 
For quantitative evaluation, we adopt the norm of face embeddings as the quality metric, following MagFace~\cite{meng2021magface}. As reported in Table~\ref{tab:main_gen}, InsightAR achieves the highest MagFace score among autoregressive models with the same number of tokens per image. Figure \ref{fig:face_text_gen} further provides a comparison between LlamaGenTok-AR and InsightAR, illustrating that InsightTok’s improved face reconstruction consistently translates into higher-quality face generation.

\textbf{Text rendering.} We evaluate text rendering performance by prompting the model to generate long-form paragraphs on blank backgrounds (examples shown in the right part of Figure~\ref{fig:face_text_gen} and Appendix~\ref{app:gen_text}). An OCR model~\cite{trullemans2016doctr} is used to recognize the rendered text, and normalized edit distance (T-NED) is computed against the ground truth. As shown in Table~\ref{tab:main_gen} and Figure~\ref{fig:face_text_gen}, InsightAR consistently generates long-form text with higher accuracy, demonstrating that improved text reconstruction is a key prerequisite for faithful text rendering.

\textbf{General text-to-image generation.} We further evaluate InsightAR on standard text-to-image benchmarks~\cite{geneval,dpg}. As reported in Table~\ref{tab:main_gen}, InsightAR achieves performance comparable to Janus-Pro~\cite{janus} and other autoregressive image models on general multimodal generation tasks. Figure~\ref{fig:gen_vis} presents qualitative comparisons between InsightAR and Janus-Pro using the same prompts (listed in Appendix~\ref{app:gen_vis}), where InsightAR consistently produces images with stronger photorealism, clearer text, and more faithful facial details. These results indicate that our targeted enhancement of text and faces does not compromise general image generation capability. Additional visualizations are provided in Appendix~\ref{app:vis}.

\subsection{Analytic Experiments}

\definecolor{darkgreen}{RGB}{0,120,0}
\definecolor{darkred}{RGB}{150,0,0}

\newcommand{\cmark}{\ding{51}} % ✓
\newcommand{\xmark}{\ding{55}} % ✗

% \begin{table*}[t]
%     \centering
%     %\captionsetup{font={footnotesize}}
%     \caption{\textbf{Effect of the perceptual loss, decoder finetune and loss reweighting. }}
%     \label{tab:method_abl}
%     \resizebox{0.85\linewidth}{!}{
%     \begin{tabular}{ccc|llll}
%     \toprule
%         $\mathcal{L}_{\text{text}}\,\&\, \mathcal{L}_{\text{face}}$ & Loss Reweighting  & Decoder Finetune &  T-ACC$_m\uparrow$ & Face-Sim$_m\uparrow$ & rFID $\downarrow$ & IN-PSNR $\uparrow$ \\
%         \midrule
%         \xmark & -& -& 30.89& 0.29 & 0.60& 23.65 \\
%         \cmark & \xmark& \xmark& 55.18 \textcolor{darkgreen}{($\uparrow$ 24.29)}
%  & 0.42 \textcolor{darkgreen}{($\uparrow$ 0.13)}
% & 1.11 \textcolor{darkred}{($\uparrow$ 0.51)} & 22.41 \textcolor{darkred}{($\downarrow$ 1.24)}\\
%          \cmark&\cmark & \xmark& 51.12 \textcolor{darkgreen}{($\uparrow$ 20.23)}& 0.36 \textcolor{darkgreen}{($\uparrow$ 0.07)}& 0.64& 23.32\\
%          \cmark & \cmark& \cmark& 53.05 \textcolor{darkgreen}{($\uparrow$ 22.16)}& 0.36 \textcolor{darkgreen}{($\uparrow$ 0.07)}& 0.69 & 23.64 \\
%         \bottomrule
%     \end{tabular}
%     }
% \end{table*}

\begin{table*}[t]
    \centering
    \vspace{-10pt}
    %\captionsetup{font={footnotesize}}
    \caption{\textbf{Effect of specialized perceptual losses and area-based loss weighting} (Section~\ref{sec:text}).}
    \label{tab:method_abl}
    \vspace{-3pt}
    \resizebox{0.95\linewidth}{!}{
    \begin{tabular}{cc|llll}
    \toprule
        $\mathcal{L}_{\text{text}}\,\&\, \mathcal{L}_{\text{face}}$ & Area-based Weighting &  T-ACC$_m\uparrow$ & Face-Sim$_m\uparrow$ & rFID $\downarrow$ & IN-PSNR $\uparrow$ \\
        \midrule
        \xmark & - & 30.89& 0.29 & 0.60& 23.65 \\
        \cmark & \xmark& 55.18 \textcolor{darkgreen}{($\uparrow$ 24.29)}
 & 0.42 \textcolor{darkgreen}{($\uparrow$ 0.13)}
& 1.11 \textcolor{darkred}{($\uparrow$ 0.51)} & 22.41 \textcolor{darkred}{($\downarrow$ 1.24)}\\
         \cmark & \cmark& 53.05 \textcolor{darkgreen}{($\uparrow$ 22.16)}& 0.36 \textcolor{darkgreen}{($\uparrow$ 0.07)}& 0.69 \textcolor{darkred}{($\uparrow$ 0.09)} & 23.64 \textcolor{darkred}{($\downarrow$ 0.01)}\\
        \bottomrule
    \end{tabular}
    }
    \vspace{-5pt}
\end{table*}
\begin{table}[t]

    \centering
    %\captionsetup{font={footnotesize}}
    \caption{\textbf{Training only the decoder yields minimal gains.} (Experiments in a smaller-scale setting)}
    \label{tab:discuss_abl}
    \resizebox{0.65\linewidth}{!}{
    \begin{tabular}{l|ccc}
    \toprule
    Method &  T-ACC$_m$ & T-NED$_m$ & Face-Sim$_m$ \\
        \midrule
    Vanilla VQGAN& 23.52 & 43.59 & 0.27  \\
    + $\mathcal{L}_{\text{text}}\,\&\, \mathcal{L}_{\text{face}}$ on decoder only & 24.15 & 43.92 & 0.28\\
    + $\mathcal{L}_{\text{text}}\,\&\, \mathcal{L}_{\text{face}}$ on full model & \textbf{40.64} & \textbf{59.68} & \textbf{0.35} \\
        \bottomrule
    \end{tabular}
    }
\vspace{-5pt}
\end{table}

\textbf{Effect of specialized perceptual losses} is ablated in Table \ref{tab:method_abl}. Compared to the baseline model, incorporating text and face perceptual losses yields substantial improvements in text and face reconstruction. However, without area-based loss weighting, these specialized losses dominate optimization and noticeably degrade reconstruction quality in other regions, as evidenced by worse rFID and PSNR. In contrast, our proposed reweighting scheme effectively balances domain-specific gains with overall reconstruction quality, resulting in only minor changes to rFID and PSNR.

\textbf{Are we only enhancing the decoder?} As shown in Table~\ref{tab:discuss_abl}, applying the specialized perceptual losses only to the decoder of a vanilla VQGAN, while freezing the encoder and quantizer, yields only marginal improvements in text and face reconstruction. This suggests that the gains from InsightTok do not stem from a stronger decoder alone, but from a \textit{refined latent representation} that better preserves fine-grained visual details.

\begin{table}[ht]
\vspace{-10pt}
\centering
\begin{minipage}[t]{0.49\linewidth}
\centering
%\captionsetup{font={footnotesize}}
\caption{\textbf{Comparison with OCR-VQGAN.}}
\label{tab:discuss_ocrvqgan}
\resizebox{\linewidth}{!}{
\begin{tabular}{l|cc}
\toprule
Method & T-ACC$_m$ & T-NED$_m$ \\
\midrule
OCR-VQGAN & 12.76 & 28.58 \\
InsightTok w/ $\mathcal L_{\text{OCR-VQGAN}}$ & 28.19 & 48.13 \\
InsightTok w/ Our $\mathcal L_{\text{text}}$ & \textbf{40.64} & \textbf{59.68} \\
\bottomrule
\end{tabular}
}
\end{minipage}
\hfill
\begin{minipage}[t]{0.5\linewidth}
\centering
%\captionsetup{font={footnotesize}}
% \caption{InsightTok with an increased codebook size.}
\caption{\textbf{Larger codebook sizes.}}
\label{tab:codebook_abl}
\resizebox{\linewidth}{!}{
\begin{tabular}{c|c|cc}
\toprule
Codebook  & $\mathcal{L}_{\text{text}}\,\&\,\mathcal{L}_{\text{face}}$ & T-ACC$_m$ & Face-Sim$_m$ \\
\midrule
\multirow{2}{*}{16,384}  & \xmark & 30.87 & 0.29 \\
&  \cmark & \textbf{53.05} & \textbf{0.36} \\
\midrule
\multirow{2}{*}{65,536} & \xmark & 35.46 & 0.31 \\
&  \cmark & \textbf{55.69} & \textbf{0.40} \\
\bottomrule
\end{tabular}
}
\end{minipage}

% \vspace{-3pt}
\end{table}

% \begin{table}[ht]
%     \centering
%     %\captionsetup{font={footnotesize}}
%     \caption{Comparison with OCR-VQGAN.}
%     \label{tab:discuss_ocrvqgan}
%     \resizebox{0.5\linewidth}{!}{
%     \begin{tabular}{l|cc}
%     \toprule
%     Method &  T-ACC$_m$ & T-NED$_m$ \\
%         \midrule
%     OCR-VQGAN  & 12.76 & 28.58 \\
%     InsightTok w/ $\mathcal L_{\text{OCR-VQGAN}}$ & 28.19& 48.13 \\
%     InsightTok w/ Our $\mathcal L_{\text{text}}$ & \textbf{40.64} & \textbf{59.68}  \\
%         \bottomrule
%     \end{tabular}
%     }
%     \vspace{-3pt}
% \end{table}

% \begin{table}[ht]
%     \centering
%     %\captionsetup{font={footnotesize}}
%     \caption{InsightTok with an increased codebook size.}
%     \label{tab:codebook_abl}
%     \resizebox{0.5\linewidth}{!}{
%     \begin{tabular}{cc|c|cc}
%     \toprule
%         Codebook & BPP &  $\mathcal{L}_{\text{text}}\,\&\,\mathcal{L}_{\text{face}}$ & T-ACC$_m$ & Face-Sim$_m$ \\
%         \midrule
%          \multirow{2}{*}{16,384} & \multirow{2}{*}{0.0547} & \xmark& 30.87&0.29 \\
%          &  & \cmark &\textbf{53.05}& \textbf{0.36} \\
%         \midrule
%         \multirow{2}{*}{65,536}& \multirow{2}{*}{0.0625} & \xmark& 35.46& 0.31\\
%         & & \cmark & \textbf{55.69}& \textbf{0.40} \\
%         \bottomrule
%     \end{tabular}
%     }
%         \vspace{-3pt}
% \end{table}

\textbf{Comparison with OCR-VQGAN}~\cite{ocrvqgan} is shown in Table~\ref{tab:discuss_ocrvqgan}. Overall, OCR-VQGAN significantly lags behind InsightTok in text reconstruction quality. To further investigate, we conducted a controlled experiment by replacing our proposed text perceptual loss $\mathcal L_{\text{text}}$ with the perceptual loss $\mathcal L_{\text{OCR-VQGAN}}$ proposed by OCR-VQGAN~\cite{ocrvqgan}. We found that $\mathcal L_{\text{OCR-VQGAN}}$ is less effective than our approach, likely because its global supervision is less sensitive to text patterns than our localized loss.
% likely because it applies a global loss that is less sensitive to the text patterns than our localized approach.

\textbf{Larger codebook sizes.} As shown in Table~\ref{tab:codebook_abl}, our method consistently improves performance with both 16k and 65k codebooks, indicating that the proposed framework scales effectively to tokenizers with larger bottleneck capacities.

\textbf{Additional analytic experiments} on model size, detector recall rate, the individual effects of $\mathcal L_{\text{text}}$ and $\mathcal L_{\text{face}}$, and comparisons with continuous tokenizers are provided in Appendix~\ref{sec:extra_abl}.

\textbf{Extra overhead.} Our training framework adds \textit{around 2\% overhead} compared to vanilla VQGAN; detailed profiling is provided in Appendix~\ref{sec:overhead}. Note that text and face detection are performed \textit{offline} during data preprocessing and are not part of the training loop.

% Limitations: Currently only focus on English text, other language is not specifically optimized. Use mixed precision, may have rounding errors. Does not perform quality filtering. Data scale is limited.

% \textbf{Impact of Model Size} Table \ref{tab:size_abl}

\section{Conclusion}
% In this paper, we identified an importance misalignment between standard tokenizer objectives and human perceptual priorities, and proposed the InsightTok framework to address this issue. 

In this paper, we attribute the challenges of discrete image generation in text- and face-centric scenarios to the lack of targeted supervision in image tokenizer training.
To address this, we propose InsightTok, which introduces localized, domain-specific perceptual losses to enhance text readability and facial fidelity.
Extensive experiments demonstrate that InsightTok outperforms existing tokenizers by a large margin on text and face reconstruction, while maintaining competitive general reconstruction quality.
These improvements consistently transfer to downstream autoregressive image generation, resulting in higher-quality text and face synthesis in both quantitative evaluations and qualitative comparisons.
Our findings suggest that aligning tokenizer supervision with perceptually critical content is a practical and effective way to improve discrete image generation.

% spatially adaptive

% \begin{ack}
% Use unnumbered first level headings for the acknowledgments. All acknowledgments
% go at the end of the paper before the list of references. Moreover, you are required to declare
% funding (financial activities supporting the submitted work) and competing interests (related financial activities outside the submitted work).
% More information about this disclosure can be found at: \url{https://neurips.cc/Conferences/2026/PaperInformation/FundingDisclosure}.

% Do {\bf not} include this section in the anonymized submission, only in the final paper. You can use the \texttt{ack} environment provided in the style file to automatically hide this section in the anonymized submission.
% \end{ack}

{
\small

\bibliographystyle{plain}
\bibliography{refs}
}

%%%%%%%%%%%%%%%%%%%%%%%%%%%%%%%%%%%%%%%%%%%%%%%%%%%%%%%%%%%%

\appendix

\section{Limitations and Broader Impact}
\label{app:limitation}
\textbf{Limitations.} The proposed approach is designed to enhance reconstruction quality in the targeted domains, namely text and faces, and is not intended to improve reconstruction across all types of visual content. In addition, our current treatment of text in both reconstruction and generation is limited to English text and English prompts. Extending InsightTok to other languages follows the same general principle, and we leave this direction to future work.

\textbf{Broader impact and risks.} On the positive side, improved text and face fidelity may benefit accessibility, design, education, and creative workflows that require accurate visual rendering. However, improved facial fidelity also introduces dual-use risks, including identity cloning and impersonation (e.g., deepfakes). To mitigate these risks, we restrict our models to non-commercial academic research and avoid using personal names or explicit identity labels during tokenizer training. We further recommend that downstream systems adopt safeguards such as watermarking and monitoring mechanisms. In addition, the use of pretrained face recognition models may propagate demographic biases present in their training data; therefore, practitioners are encouraged to use more diverse and better-calibrated recognition models, and to audit performance across demographic groups.

\section{Formulation: VQ-EMA with Random Restart}
\label{app:quant}
InsightTok adopts a VQ-EMA quantizer with random codebook restart, with a codebook size of 16,384 and an embedding dimension of 256. This choice avoids the additional information loss incurred by projecting latents to very low dimensions prior to quantization~\cite{vitvqgan}. In practice, despite operating in a relatively high-dimensional latent space ($d=256$), codebook utilization quickly rises to near 100\% early in training and remains high thereafter.

\textbf{EMA codebook update.}
Formally, let the codebook be $\{\bm e_k\in \mathbb R^d\}_{k=1}^K$. We maintain two running statistics for each code:
\begin{itemize}[itemsep=0pt, topsep=0pt]
\item Cluster sum $S_k\in \mathbb R^d$: the running sum  of features assigned to code $k$.
\item Cluster size $N_k\in\mathbb R$: the running count of assignments to code $k$.
\end{itemize}
Each codebook entry is computed as the centroid of its associated cluster:
\begin{equation}
\bm e_k := \frac{S_k} {N_k}.
\end{equation}
At each training iteration, the encoder produces a batch of latent vectors $\{\bm z^{(i)}\in\mathbb R^d\}_{i=1}^{M}$ (aggregated over all images in the batch). Each latent is assigned to its nearest codeword:
\begin{equation}
\hat{\bm z}^{(i)}=\bm e_k \iff \bm z^{(i)} \in \mathcal C_k \iff k=\arg\min_{k\in[K]} \|\bm z^{(i)} - \bm e_k \|^2,
\end{equation}
where $\hat{\bm z}^{(i)}$ denotes the quantized embedding of $\bm z^{(i)}$, and $\mathcal C_k$ is the set of latents assigned to code $k$ at the current iteration. Given the partition $\{\mathcal C_k\}_{k=1}^K$, we update the cluster statistics with exponential moving average:
\begin{equation}
\begin{aligned}
S_k^{\text{new}} &\leftarrow (1-\mu) S_k^{\text{old}} + \mu \sum_{\bm z^{(i)}\in \mathcal C_k}\bm z^{(i)}, \\
N_k^{\text{new}} &\leftarrow (1-\mu) N_k^{\text{old}} + \mu |\mathcal C_k|,
\end{aligned}
\end{equation}
where $\mu$ is the EMA update ratio. Since the codebook embeddings are updated without direct gradient-based optimization, the codebook loss serves only to constrain the encoder outputs to remain close to the distribution of the selected code embeddings:
\begin{equation}
\mathcal L_{\text{codebook}}=\big\|\bm z-\operatorname{sg}(\hat{\bm z})\big\|_2^2,
\end{equation}
where $\operatorname{sg}(\cdot)$ denotes the stop-gradient operator, and $\bm z$ and $\hat{\bm z}$ represent the latent features before and after quantization, respectively.

\paragraph{Random restart for dead codes.}
Over training, some codewords may drift away from the latent distribution and become rarely selected (so-called \emph{dead} codes). To maintain high utilization, we periodically reinitialize dead codes using randomly sampled encoder outputs. Concretely, we monitor the running usage statistic $N_k$ and reinitialize code $k$ whenever $N_k$ falls below $1$. This simple restart mechanism helps keep the codebook well-covered throughout training.

\section{Additional Analytic Experiments}
\label{sec:extra_abl}
\textbf{Scaling model size.} Table~\ref{tab:size_abl} shows that reconstruction performance improves as model size increases.
\begin{table}[ht]
    \centering
    \vspace{-15pt}
    %\captionsetup{font={footnotesize}}
    \caption{Effect of scaling up the model size.}
    \label{tab:size_abl}
    \resizebox{0.65\linewidth}{!}{
    \begin{tabular}{ccc|ccc}
    \toprule
      Base Channels &  \#Res-Blocks & \#Params &  T-ACC$_m$ & F-Sim$_m$ \\
        \midrule
     128 &  2 & 72M & 37.91 & 0.32 \\
    128 &  4& 111M&40.64 & 0.35\\
     256 &  4 & 426M & 53.05& 0.36 \\
        \bottomrule
    \end{tabular}
    }
\end{table}

\textbf{Isolating text and face supervision.} Table~\ref{tab:isolate} shows that text-aware and face-aware losses specifically improve their corresponding targets. Combining both losses achieves strong performance with only a slight trade-off compared to optimizing each individually.

\begin{table}[ht]
    \centering
    \vspace{-10pt}
    %\captionsetup{font={footnotesize}}
    \caption{Isolating text and face supervision.}
    \label{tab:isolate}
    \resizebox{0.42\linewidth}{!}{
    \begin{tabular}{cc|cc}
    \toprule
    $\mathcal L_{\text{text}}$ & $\mathcal L_{\text{face}}$ & T-ACC$_m$ & Face-Sim$_m$ \\
        \midrule
  \xmark & \xmark & 23.52 & 0.27 \\
  \cmark & \xmark & 41.36& 0.27 \\
  \xmark & \cmark &  23.47 & 0.37 \\
  \cmark & \cmark & 40.64	& 0.35 \\
        \bottomrule
    \end{tabular}
    }
\end{table}

\textbf{Detector coverage.} We simulate imperfect detectors with lower recall by randomly dropping portions of detected face and text regions during training. Table~\ref{tab:drop_abl} shows that better region identification leads to stronger reconstruction quality.

\begin{table}[ht]
\vspace{-10pt}
    \centering
    %\captionsetup{font={footnotesize}}
    \caption{Impact of detector coverage.}
    \label{tab:drop_abl}
    \resizebox{0.43\linewidth}{!}{
    \begin{tabular}{c|cc}
    \toprule
      Drop portion & T-ACC$_m$ & Face-Sim$_m$ \\
        \midrule
100\%&	23.52&	0.27\\
90\%&	30.08&	0.30\\
75\%&	34.03&	0.32\\
50\%&	37.37&	0.34\\
25\%&	39.39&	0.35\\
0\%&	40.64&	0.35\\
        \bottomrule
    \end{tabular}
    }
\end{table}

\textbf{Comparison with continuous tokenizers} is presented in Table~\ref{tab:cont_vae}. Continuous tokenizers provide a useful reference as an approximate upper bound, since they typically operate at much higher bottleneck capacity (e.g., Bit-Per-Pixel $\geq$ 1) and thus achieve stronger reconstruction quality. That said, discrete and continuous tokenizers are designed for different generative interfaces, namely autoregressive modeling and diffusion modeling, respectively, and are therefore not directly interchangeable.

\begin{table}[ht]
    \centering
    \vspace{-5pt}
    %\captionsetup{font={footnotesize}}
    \caption{Comparison with continuous tokenizers. For continuous representations, we assume each channel is encoded using 32 bits}
    \label{tab:cont_vae}
    \resizebox{0.7\linewidth}{!}{
    \begin{tabular}{c|cccc}
    \toprule
Method	&Bit-Per-Pixel&	Downstream Model&	T-ACC$_m$ & Face-Sim$_m$\\
\midrule
InsightTok&	0.0547&	Autoregressive&	53.05&	0.36\\
DC-AE~\cite{dcae}&	1.0000&	Diffusion&	38.25&	0.39\\
SDXL-VAE~\cite{sdxl}&	2.0000&	Diffusion&	56.86&	0.55\\
FLUX-VAE~\cite{flux2024}&	8.0000&	Diffusion&	87.64&	0.86\\
        \bottomrule
    \end{tabular}
    }
\end{table}

\section{Extra Overhead}
\label{sec:overhead}
Text and face detection (bounding boxes and landmarks) are performed offline during data preprocessing and are not part of the training loop. During training, only small cropped regions are passed to the recognition models for perceptual supervision, contributing minimal overhead. As shown in Table~\ref{tab:overhead_compute}, the added recognition models account for less than 2\% of total FLOPs. We assume each image contains, on average, 1 facial region and 2 textual regions.

We further report the training latency comparison in Table~\ref{tab:overhead_latency}. Under the same setup (batch size 16, averaged over 10 iterations), the training time per iteration increases only slightly from 2056 ms to 2099 ms per iteration (approximately 2\%), while GPU memory usage remains almost unchanged (44.1 GB → 44.3 GB).

Overall, both theoretical (FLOPs) and empirical (wall-clock) measurements show that our method adds only minimal overhead, while delivering substantial improvements in reconstruction quality.

\begin{table}[ht]
    \centering
    %\captionsetup{font={footnotesize}}
    \caption{Theoretical computational cost of each component in InsightTok.}
    \label{tab:overhead_compute}
    \resizebox{0.9\linewidth}{!}{
    \begin{tabular}{l|cccc}
    \toprule
        Component&	\#Instances$\times$Resolution&	Model Size&	GFLOPs&	Percentage (\%) \\
        \midrule
         Autoencoder &	1$\times$256$\times$256&	426M&	2575&	98.32 \\
Discriminator&	1$\times$256$\times$256	&2.8M	&6.3&	0.24 \\
Face Recognition Model&	1$\times$112$\times$112&	43.6M&	12.6&	0.48\\
Text Recognition Model&	2$\times$32$\times$128&	23.5M&	25.0&	0.95\\
        \bottomrule
    \end{tabular}
    }
\end{table}

\begin{table}[ht]
    \centering
    %\captionsetup{font={footnotesize}}
    \caption{Per-stage latency and GPU memory footprint during a training iteration. Latency is measured in milliseconds.}
    \label{tab:overhead_latency}
    \resizebox{1.0\linewidth}{!}{
    \setlength{\tabcolsep}{2pt}
    \begin{tabular}{l|ccccc|c|c|c}
    \toprule
    \multirow{2}{*}{\textbf{Method}} & \multicolumn{5}{c|}{\textbf{Forward}} & \multirow{2}{*}{\textbf{Backward}} & \multirow{2}{*}{\textbf{Total Time}} & \multirow{2}{*}{\textbf{GPU Memory}} \\
&	Autoencoder&	Face Warping&	Face Model&	Text Model&	Discriminator& & &\\
\midrule
Vanilla VQGAN&	581.42±1.82&	-&	-&	-&	4.61±0.06 &	1470.27±1.97&	2056.3&	44066±34 \\
InsightTok&	578.56±2.38&	3.19±0.20&	14.86±0.41&	21.59±2.74&	4.37±0.08&	1476.93±2.86&	2099.5&	44333±35\\
        \bottomrule
    \end{tabular}
    }
\end{table}

\section{Implementation Details}
\label{app:impl}
\subsection{InsightTok}
\label{app:impl_tok}
\textbf{Architecture.} InsightTok follows the convolutional VQGAN architecture~\cite{vqgan}. It consists of five convolutional stages with an overall downsampling factor of $16\times$. The base channel width is 256, and we use 4 residual blocks per stage in both the encoder and decoder. We adopt a codebook of size 16,384 with embedding dimension 256; codebook embeddings are updated with EMA, and rarely used codes are periodically reinitialized when their usage drops below a threshold. The resulting tokenizer contains 426M parameters.

\textbf{Training data.} We employ two types of data for tokenizer training:
\begin{itemize}[itemsep=0pt, topsep=0pt]
\item Diverse, unannotated data: A large-scale dataset of over 100M images collected from ImageNet~\cite{imagenet} and LAION~\cite{laion}, used to support general-purpose visual reconstruction.
\item Region-annotated data: Text- and face-containing subsets of LAION~\cite{chen2023textdiffuser,zheng2022general}, where each image is associated with localization annotations, including text bounding boxes, face bounding boxes, and facial landmarks. Annotations are obtained using a text detector~\cite{db} and a face detector~\cite{deng2019arcface}. The resulting datasets contain approximately 4M images with text annotations and 8M images with face annotations.
\end{itemize}

\textbf{Hyperparameters.} We use an $\ell_1$ reconstruction loss $\mathcal L_{\text{rec}}=\|\bm x-\hat{\bm x}\|_1$, a codebook optimization loss $\mathcal L_{\text{codebook}}$ with weight $\beta=0.02$, an LPIPS perceptual loss~\cite{lpips} with weight $\gamma=0.5$, and an adversarial loss $\mathcal L_{\text{GAN}}$ with a PatchGAN~\cite{patchgan} discriminator and weight $\eta=0.5$. Following MaskBiT~\cite{maskbit}, we additionally apply LeCAM regularization~\cite{tseng2021regularizing} with weight $0.05$. The domain-specific perceptual losses are incorporated with default weights $\alpha_1=\alpha_2=1.0$. The overall training objective is
\begin{equation}
    \mathcal L_{\text{InsightTok}}
=\mathcal L_{\text{rec}}
+\beta\cdot \mathcal L_{\text{codebook}}
+\gamma\cdot \mathcal L_{\text{perc}}
+\eta\cdot \mathcal L_{\text{GAN}} + \alpha_1 \cdot \mathcal L_{\text{text}} + \alpha_2 \cdot \mathcal L_{\text{face}}.
\end{equation}
We train the tokenizer on $256^2$ images with a batch size of 512. Optimization is performed using Adam~\cite{adam2014method} with $\beta_1=0.5$ and $\beta_2=0.9$, gradient clipping set to 1.0, and weight decay of 0.05. The learning rate is $1\times10^{-4}$, following a cosine decay schedule with a linear warmup over the first 5\% of training steps.

\textbf{Training procedure.} We adopt a three-phase training procedure:
\begin{itemize}[itemsep=0pt, topsep=0pt]
\item Pretraining: the tokenizer is trained for 200k steps on large-scale unannotated data using the standard VQGAN objective $\mathcal L_{\text{image}}$.
\item Text and Face Training: we enable the proposed text and face perceptual losses $\mathcal L_{\text{text}},\mathcal L_{\text{face}}$ and continue training for an additional 40k steps on a mixture of region-annotated data and a small portion of unannotated images.
\item Decoder Fine-tuning: we freeze the encoder and quantizer and train only the decoder for 40k steps, reducing the loss weights to $\alpha_1=\alpha_2=0.1$ to further refine reconstruction quality.
\end{itemize}
Full training is conducted on 32 A100 GPUs for approximately 5 days. For ablation studies reported in Table~\ref{tab:discuss_abl},~\ref{tab:discuss_ocrvqgan},~\ref{tab:isolate},~\ref{tab:drop_abl}, we adopt a \textit{lightweight setup} using a smaller model with 111M parameters, trained for 100k steps over the first two phases only, without decoder fine-tuning.

\subsection{InsightAR}
\label{app:impl_ar}
\textbf{Architecture.} The architecture of InsightAR largely follows the Janus-Pro~\cite{janus} framework. An MLP adapter maps visual codebook embeddings into the input embedding space of a 7B-parameter large language model.
% The transformer is initialized from Janus-Pro pretrained weights, while the adapter and visual token prediction head are randomly initialized.
During training and inference, the sequence of discrete image tokens is concatenated with the input text prompt tokens, separated by a dedicated start-of-image token.

\textbf{Training data.}
We train InsightAR on a large-scale collection of publicly available text-to-image datasets, including LAION~\cite{laion}, Flux-Reason-6M~\cite{fang2025flux}, and Echo-4o~\cite{ye2025echo}. The data are filtered using the LAION aesthetic predictor\footnote{\href{https://github.com/LAION-AI/aesthetic-predictor}{https://github.com/LAION-AI/aesthetic-predictor}} and resolution-based criteria to ensure image quality. To further evaluate long-form text rendering, we augment the training set with synthetic text rendering data generated using an off-the-shelf tool\footnote{We use the tools provided by \href{https://github.com/GbotHQ/ocr-dataset-rendering}{https://github.com/GbotHQ/ocr-dataset-rendering}}. 
The final training corpus contains approximately 150M images. All images are transformed to a resolution of $512^2$, corresponding to sequences of 1,024 discrete image tokens.

\textbf{Training procedure.} Following the training strategy of Janus-Pro~\cite{janus}, we adopt a staged training process that first warms up the randomly initialized components before training the full model. Optimization is performed using AdamW~\cite{adamw} with $\beta_1=0.9$ and $\beta_2=0.95$, zero weight decay, and gradient clipping set to 1.0. The batch size is fixed to 512.
%Specifically, training proceeds in two stages:
Training proceeds in two stages:
\begin{itemize}[itemsep=0pt, topsep=0pt]
    \item Stage 1: the language model is frozen, and only the MLP adapter and the visual token prediction head are trained. We use a learning rate of $1\times10^{-3}$ and train for 20k steps on a small subset of LAION.
    \item Stage 2: we train the full model with a constant learning rate of $1\times10^{-4}$ on the full large-scale dataset for one epoch.
\end{itemize}
In addition, we conduct a controlled experiment by training an autoregressive model with the LlamaGen tokenizer~\cite{llamagen}, which is used in the original Janus-Pro framework, following the same training recipe. This variant is denoted as \textit{LlamaGenTok-AR}.

\section{Benchmark and Evaluation Protocol}
\subsection{Image Reconstruction}
\label{app:recon}
\subsubsection{Compression Rate}
The compression rate of a visual tokenizer is measured in bits per pixel (BPP), defined as the total number of bits used to represent an image divided by its spatial resolution. The total number of bits is computed as the product of the number of tokens per image and the information capacity of the codebook. Formally,
\begin{equation}
\text{BPP} = \frac{\text{Tokens-Per-Image}\times\log_2(\text{Codebook-Size})}{H\times W},
\end{equation}
where $(H, W)$ denotes the spatial resolution of the image.

\subsubsection{Text Reconstruction}
Text reconstruction is evaluated on TokBench~\cite{tokbench}, which consists of text-centric images with diverse fonts, styles, scales, and backgrounds. Each image is annotated with text locations and corresponding transcriptions, which serve as ground truth. TokBench adopts the PARSeq~\cite{bautista2022scene} text recognizer to evaluate reconstructed images, reporting text recognition accuracy (T-ACC) and normalized edit distance (T-NED) as evaluation metrics. Specifically, T-ACC measures word-level accuracy and is positive only when the recognized text exactly matches the ground truth, while T-NED provides a more fine-grained measure of character-level similarity~\cite{tuo2023anytext}:
\begin{equation}
\label{eq:ned}
\text{T-NED}=1-\frac{D(s,\hat{s})}{\max(l,\hat{l})},
\end{equation}
where $s$ and $\hat{s}$ denote the recognized text and ground-truth transcription, $l$ and $\hat{l}$ are their corresponding character lengths, and $D(\cdot,\cdot)$ is the edit distance. Text instances are further grouped into \emph{small}, \emph{medium}, and \emph{large} categories based on their spatial size. Metrics with the subscript ``$s$'' (T-ACC$_s$ and T-NED$_s$) are averaged over small instances, while metrics with the subscript ``$m$'' (T-ACC$_m$ and T-NED$_m$) are averaged over all instances across the three groups.

\subsubsection{Face Reconstruction}
Face reconstruction is also evaluated on TokBench~\cite{tokbench}, which contains faces captured in natural, unconstrained environments, with multiple faces appearing in each image. 
The benchmark leverages an off-the-shelf face recognition model~\cite{deng2019arcface} to extract face embeddings from both the ground-truth and reconstructed images. Face quality is measured using the cosine similarity between the corresponding embeddings, reported as the face similarity metric (F-Sim).
Face instances are grouped into \emph{small}, \emph{medium}, and \emph{large} categories based on their spatial size. Metrics with the subscript ``$s$'' (F-Sim$_s$) are averaged over small instances, while metrics with the subscript ``$m$'' (F-Sim$_m$) are averaged over all instances across the three groups.

\subsubsection{General visual reconstruction}
For overall reconstruction quality, we report reconstruction FID (rFID) and peak signal-to-noise ratio (PSNR). Both metrics are computed on 50k images from the validation set of ImageNet-1K~\cite{imagenet}.

\subsection{Image Generation}
\label{app:gen}
All images in this study are generated using classifier-free guidance with a guidance scale of 5.0 and top-k sampling with k=4096.

\subsubsection{Face Generation}
\label{app:gen_face}
\begin{figure}[h]
    \centering
    \includegraphics[width=0.8\linewidth]{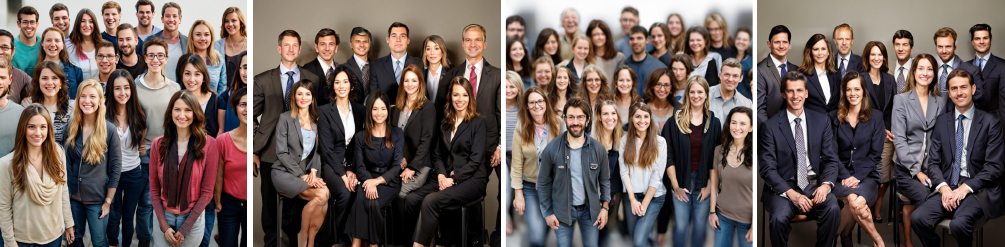}
    \caption{Additional face generation examples produced by InsightAR.}
    \label{fig:face_extra}
\end{figure}

We evaluate face generation quality in a challenging crowd-generation setting, where models are prompted to synthesize images containing multiple individuals (example prompts are shown below). In this scenario, each face occupies a relatively small region of the image, making the task particularly sensitive to the tokenizer’s ability to preserve fine-grained facial details.
We construct 15 prompts covering diverse configurations, including varying numbers of people (up to twenty), poses (sitting or standing), and clothing styles (casual or professional). Additional qualitative examples are provided in Figure~\ref{fig:face_extra}.
For quantitative evaluation, we first detect all faces in the generated images using an off-the-shelf face detector~\cite{deng2019arcface}. We then compute the norm of the corresponding face embeddings as a quality metric, following MagFace~\cite{meng2021magface}. The reported scores are averaged over 300 generated images for each compared model.

\begin{tcolorbox}[promptbox={Prompt Example: Face Generation}]
Ten people stand together closely in a well-lit, neutral indoor setting, forming a group portrait that captures a sense of camaraderie and diverse individuality. The group consists of five men and five women, arranged in two rows: six standing in the back and four seated or crouching slightly in the front, creating a balanced composition. Their ages range from early twenties to mid-forties, with a variety of ethnic backgrounds reflected in skin tones, facial features, and hair types. Each person wears contemporary casual clothing in varied but complementary colors, including denim jeans, sweaters, button-up shirts, and light jackets, with textures visible such as soft cotton, knit wool, and smooth denim. The individuals display relaxed, natural expressions mostly soft smiles or neutral gazes directly engaging the camera.
\end{tcolorbox}

\subsubsection{Text Rendering}
\label{app:gen_text}
\begin{figure}[h]
    \centering
    \includegraphics[width=0.8\linewidth]{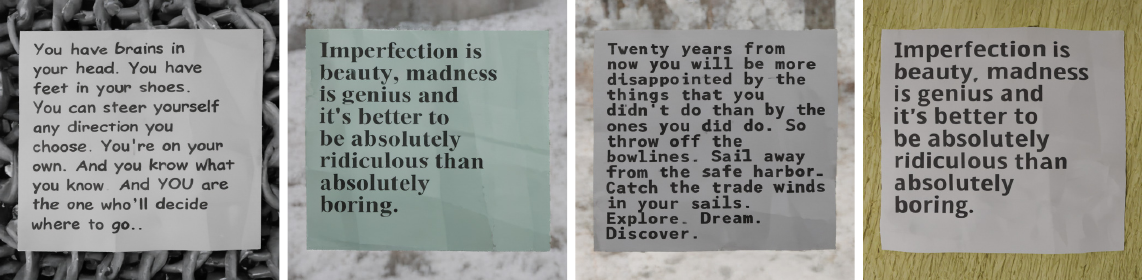}
    \caption{More text images generated by InsightAR.}
    \label{fig:text_extra}
\end{figure}

We evaluate the model’s text rendering ability by prompting it to generate long-form paragraphs on blank backgrounds, with an example prompt shown below. An OCR model~\cite{trullemans2016doctr} is used to transcribe the generated text, and we compute the normalized edit distance (Eq.~\ref{eq:ned}) against the ground truth. We sample 200 English quotes\footnote{\href{https://huggingface.co/datasets/Abirate/english\_quotes}{https://huggingface.co/datasets/Abirate/english\_quotes}} with lengths ranging from 100 to 300 characters, and all reported metrics are computed over this set. Additional generation examples are shown in Figure~\ref{fig:text_extra}.

\begin{tcolorbox}[promptbox={Prompt Example: Text Rendering}]
The image shows bold, clear text on a white background that reads: ``You have brains in your head. You have feet in your shoes. You can steer yourself any direction you choose. You're on your own. And you know what you know. And YOU are the one who'll decide where to go...''
\end{tcolorbox}

\textbf{General text-to-image generation.} We evaluate InsightAR's general multimodal generation ability on standard text-to-image benchmarks, including:
\begin{itemize}[itemsep=0pt, topsep=0pt]
\item GenEval~\cite{geneval}: a compositional benchmark that measures a model’s ability to follow structured prompts involving object co-occurrence, spatial relations, counts, and colors;
\item DPG-Bench~\cite{dpg}: the Dense Prompt Graph Benchmark assesses a model’s ability to interpret and generate images from dense, complex prompts that include multiple objects, detailed attributes, and intricate relationships, intended to stress prompt-following and semantic alignment between text and image content.
\end{itemize}

\subsubsection{Visualizations}
\label{app:gen_vis}
Below, we list the text prompts used to generate the images shown in Figure~\ref{fig:gen_vis} in the main paper.

\begin{tcolorbox}[promptbox={Prompt No.1}]
Young girl in traditional Romanian costume stands confidently in a greenhouse, her face radiating pride as she holds up a clean white signboard that clearly displays the word ``INSIGHT" and ``TOK" in large, bold lettering. Dressed in a white blouse adorned with red cuffs and a colorful apron featuring intricate geometric patterns and floral designs, she exudes cultural heritage while embodying the spirit of rural life. Her long brown hair is neatly braided into two pigtails, complemented by a wide-brimmed straw hat that adds a rustic charm to her ensemble. The greenhouse setting is lush and verdant, with rows of healthy tomato plants in terracotta pots stretching into the background, their green foliage contrasting beautifully with the bright red tomatoes. The transparent roof structure allows natural light to filter through, illuminating the scene and highlighting the freshness of the produce. This image captures a harmonious blend of tradition, agriculture, and youthful enthusiasm.
\end{tcolorbox}

\begin{tcolorbox}[promptbox={Prompt No.2}]
Ten people stand together closely in a well-lit, neutral indoor setting, forming a group portrait that captures a sense of camaraderie and diverse individuality. The group consists of five men and five women, arranged in two rows: six standing in the back and four seated or crouching slightly in the front, creating a balanced composition. Their ages range from early twenties to mid-forties, with a variety of ethnic backgrounds reflected in skin tones, facial features, and hair types. Each person wears contemporary casual clothing in varied but complementary colors, including denim jeans, sweaters, button-up shirts, and light jackets, with textures visible such as soft cotton, knit wool, and smooth denim. The individuals display relaxed, natural expressions mostly soft smiles or neutral gazes directly engaging the camera.
\end{tcolorbox}

\begin{tcolorbox}[promptbox={Prompt No.3}]
An athletic boxer stands poised in the center of a brightly lit boxing ring, his muscular physique and intense focus capturing the essence of a determined competitor. Dressed in red boxing gloves and a white wrist wrap, he exudes confidence and readiness as he prepares to face his opponent in his quest to become a world champion. His trunks, prominently displaying the name ``YARDA" in bold blue letters against a white background, serve as a personal emblem of his journey from adversity to greatness. The background is a dynamic blend of vibrant stadium lights and stage equipment, casting dramatic shadows and highlighting the intensity of the moment. With his fists raised and body coiled with anticipation, He embodies the spirit of resilience and ambition, showcasing both his physical prowess and the emotional weight of his transformation from a heartbroken debt collector to a potential world champion. The composition emphasizes his central position in the ring, drawing the viewer's attention to his unwavering determination and the high stakes of the impending match.
\end{tcolorbox}

\begin{tcolorbox}[promptbox={Prompt No.4}]
Three joyful children, each radiating happiness, walk hand-in-hand along a gravel path in Bend, Oregon, their laughter and smiles capturing the essence of carefree childhood. The eldest girl, with golden hair flowing in the breeze, leads the way in a cream-colored henley shirt and rolled-up blue jeans, her sneakers adding a touch of casual comfort. In the middle, a toddler dressed in a navy-and-white checkered shirt paired with beige pants and dark sneakers beams with excitement, holding hands with both siblings. To the right, a slightly younger child, also with blonde hair, wears a light beige button-down shirt and blue jeans, complemented by brown sandals. The path they tread is flanked by lush greenery and scattered shrubs, leading into a serene, open landscape under a soft, overcast sky. The composition beautifully balances the children's playful energy against the tranquil backdrop, emphasizing a sense of unity and natural beauty. The overall style is photorealistic, with a warm, candid feel that highlights the innocence and joy of the moment.
\end{tcolorbox}

\begin{tcolorbox}[promptbox={Prompt No.5}]
Positioned prominently on a rustic wooden surface, a sleek white trucker hat captures attention with its bold and vibrant design. The hat, identified as Mountain Life Apparel's MINNEWANKA HAT, features a striking logo centered on the front panel a stylized mountain rendered in deep navy blue with accents of light teal, symbolizing peaks and snow. Below the mountain graphic, the words ``MOUNTAIN LIFE" are emblazoned in crisp, capitalized black letters, reinforcing the brand's connection to outdoor adventure and nature. The hat's construction combines a solid white front panel with a breathable mesh back, offering both style and functionality. Two small eyelets near the top enhance ventilation, while the curved brim, stitched with subtle contrast thread, adds a classic touch. The backdrop is softly blurred, showcasing a lush green environment that complements the hat's theme, suggesting an outdoor setting ideal for hiking or camping. The overall composition is clean and focused, emphasizing the hat's design and its appeal to outdoor enthusiasts.
\end{tcolorbox}

\section{Additional Visualizations}
\label{app:vis}
\begin{figure}[h]
    \centering
    \includegraphics[width=0.9\linewidth]{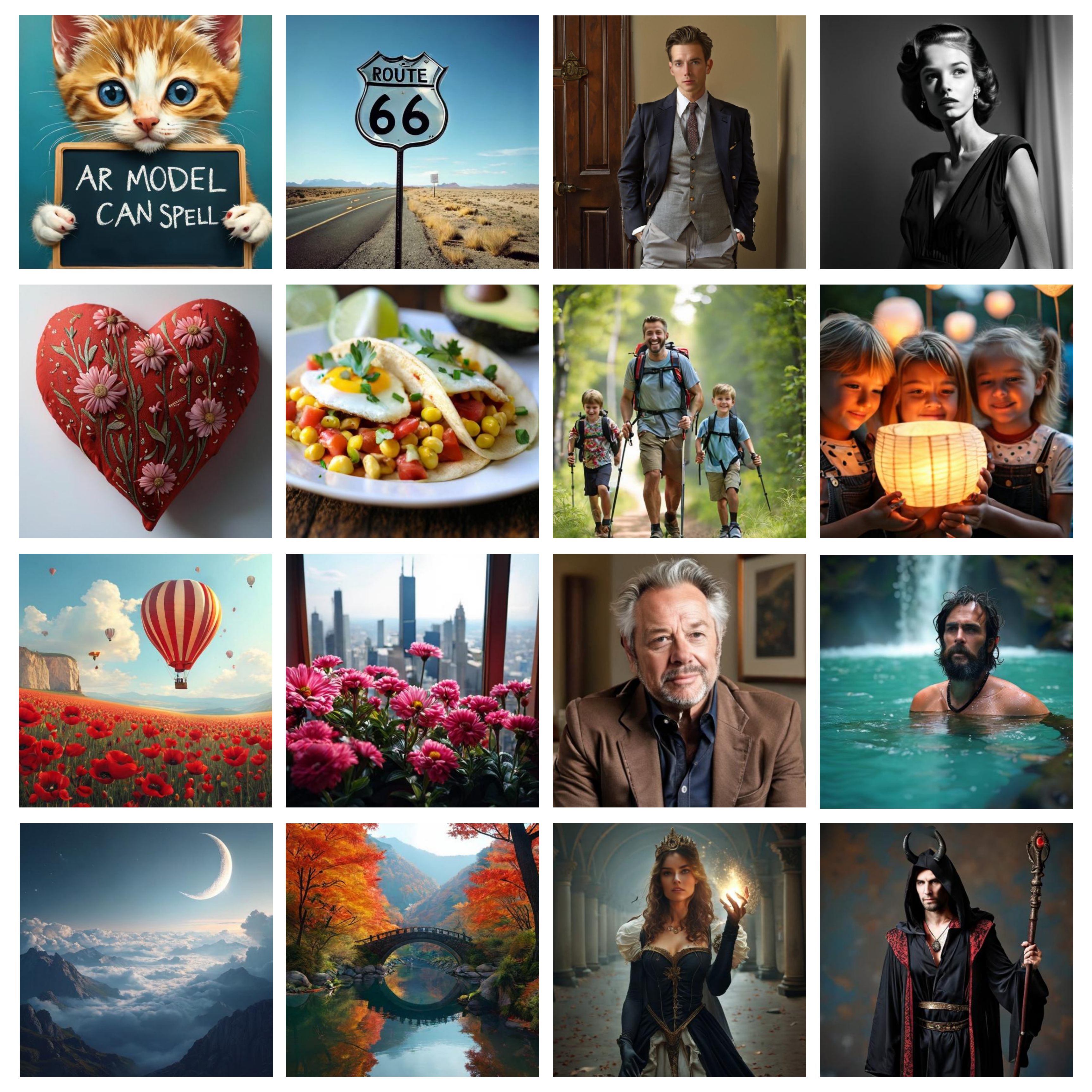}
    \caption{Qualitative examples of images generated by InsightAR.}
    \label{fig:gen_vis_extra}
\end{figure}

Figure~\ref{fig:gen_vis_extra} presents additional qualitative results for text-to-image generation using InsightAR.

% \section{Introduction to Baselines}
% \textbf{Tokenizer baselines.}

% \textbf{Autoregressive model baselines.}

% \section{Extra Visualizations}
% \subsection{Image Reconstruction}

% \subsection{Image Generation}

%%%%%%%%%%%%%%%%%%%%%%%%%%%%%%%%%%%%%%%%%%%%%%%%%%%%%%%%%%%%

% \newpage
% \input{checklist.tex}

\end{document}